\documentclass{article}

 \usepackage[preprint]{neurips_2026}

\usepackage[utf8]{inputenc} % allow utf-8 input
\usepackage[T1]{fontenc}    % use 8-bit T1 fonts
\usepackage{hyperref}       % hyperlinks
\usepackage{url}            % simple URL typesetting
\usepackage{booktabs}       % professional-quality tables
\usepackage{amsfonts}       % blackboard math symbols
\usepackage{nicefrac}       % compact symbols for 1/2, etc.
\usepackage{microtype}      % microtypography
\usepackage[table]{xcolor}         % colors

\usepackage{graphicx}
\usepackage{enumitem} %% 自定义间距
\usepackage{mathtools} %% 多行公式格式问题
\usepackage{amssymb} %% 使用数学公式中的一些命令
\usepackage{titletoc} %%% 目录
\usepackage{wrapfig}
\usepackage{xspace} %% 定义命令

\newcommand{\concat}{\mathbin{\Vert\!\Vert}}
\newcommand{\huggingface}{\raisebox{-2pt}{\includegraphics[height=1.05em]{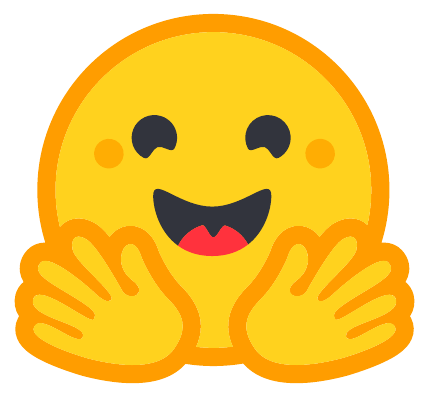}}\xspace}
\newcommand{\github}{\raisebox{-2pt}{\includegraphics[height=1.05em]{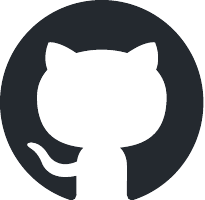}}\xspace}

\title{FactNet: A Billion-Scale Knowledge Graph for Multilingual Factual Grounding}

\author{
\textbf{
Yingli Shen$^{1*}$ \quad
Wen Lai$^{2}$\thanks{Equal contribution.} \quad
Jie Zhou$^{3}$ \quad
Xueren Zhang$^{3}$ \quad
Yudong Wang$^{1}$
} \\
\textbf{
Kangyang Luo$^{1}$ \quad
Shuo Wang$^{1}$ \quad
Ge Gao$^{4}$ \quad
Alexander Fraser$^{2}$ \quad
Maosong Sun$^{1}$\thanks{Corresponding author.} 
} \\ \\
$^1$Tsinghua University \quad
$^2$Technical University of Munich \\
$^3$ModelBest Inc. \quad
$^4$Minzu University of China \\
\texttt{syl@mail.tsinghua.edu.cn, wen.lai@tum.de}
}

\begin{document}

\maketitle

\begin{abstract}
Large language models hallucinate factual claims and struggle to ground their outputs in retrievable evidence, particularly in non-English languages.
Existing resources impose a trade-off: structured knowledge bases lack textual grounding, whereas grounded datasets remain small and monolingual.
We introduce \textbf{FactNet}, a billion-scale open resource that couples 1.7B Wikidata assertions with 3.01B evidence pointers drawn from 316 native Wikipedia editions.
FactNet employs a deterministic construction pipeline, ensuring that every evidence unit is traceable to its source with byte-level precision.
We further establish \textbf{FactNet-Bench}, an evaluation suite for Knowledge Graph Completion, Question Answering, and Fact Checking, equipped with systematic leakage controls.
Experiments demonstrate that FactNet-Bench differentiates among structural, text-aware, and LLM-integrated methods, and that cross-lingual structure enables knowledge transfer across language tiers.

\vspace{-0.5em}
\begin{center}
\renewcommand{\arraystretch}{1.2}
\footnotesize
\begin{tabular}{rll}
     \huggingface & \textbf{Dataset:} & \url{https://huggingface.co/collections/openbmb/factnet} \\
     \github & \textbf{Pipeline:} & \url{https://github.com/yl-shen/factnet} \\
\end{tabular}
\end{center}

\end{abstract}

%%% include all tex file here
\section{Introduction}
\label{sec:intro}

%% factnet structure
\begin{wrapfigure}[15]{r}{0.45\textwidth}
    \vspace{-2.5em}
    \centering
    \includegraphics[width=\linewidth]{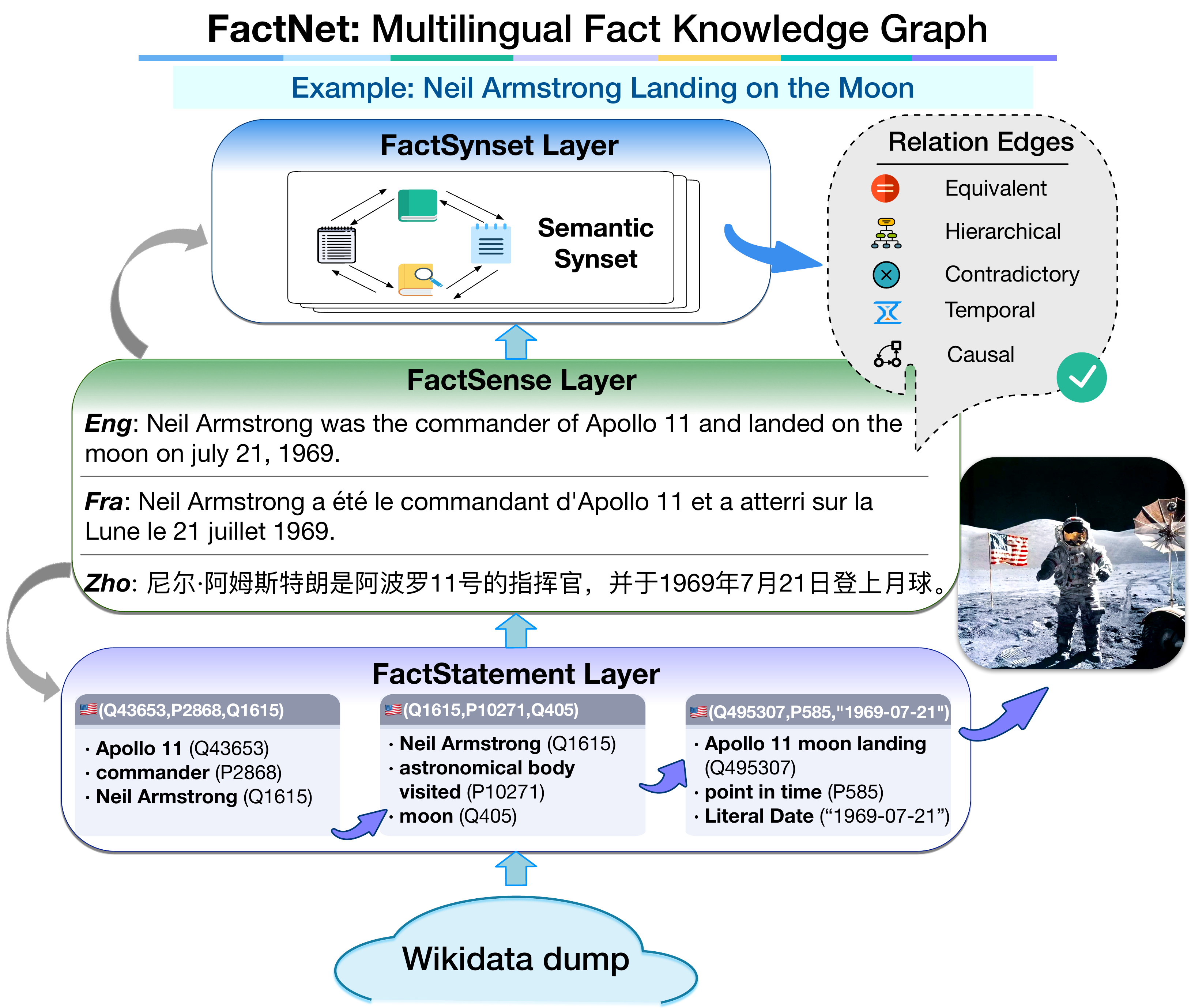}
    \vspace{-1.5em}
    \caption{\textbf{FactNet Architecture.} Three layers (\textbf{FactStatement}, \textbf{FactSense}, \textbf{FactSynset}) link Wikidata claims with Wikipedia evidence, with \textbf{RelationEdge}s enabling structural reasoning.}
    \label{fig:structure}
\end{wrapfigure}

Large Language Models (LLMs) produce fluent text yet remain unreliable in knowledge-intensive settings: they hallucinate factual claims~\citep{wang-etal-2024-factuality,huang2025survey}, conflate entities across languages~\citep{fierro-etal-2025-multilingual}, and fail to trace their outputs to verifiable sources~\citep{augenstein2024factuality}.
Retrieval-augmented generation (RAG) and knowledge-grounded systems offer a path forward, but they demand \emph{training data} that couples structured knowledge with textual evidence~\citep{gao2023retrieval,sui-etal-2025-knowledge}.
This raises a critical question: \emph{what resource can simultaneously provide structured facts, native multilingual textual evidence, and the scale required for training and evaluating modern AI systems?}

%% comparison between exist resource and factnet.
%%
\begin{table*}[t]
    \centering
    \caption{\textbf{Comparison with existing resources.} FactNet is the only resource combining billion-scale coverage and broad multilinguality with native, span-level evidence grounding. Datasets marked with $^\dagger$ also use distant supervision from Wikidata--Wikipedia alignment but lack byte-level offsets and multilingual coverage. (\textbf{Prov.}: Pipeline transparency; \textbf{MT}: Machine-translation usage.)}
    % \vspace{-0.5em}
    \label{tab:comparison_main}
    \resizebox{\textwidth}{!}{
    \begin{tabular}{@{}lccccc@{}}
    \toprule
    \textbf{Resource} & \textbf{Scale} & \textbf{Langs} & \textbf{Evidence} & \textbf{Construction Method} & \textbf{Prov.} \\ \midrule
    \rowcolor{gray!10} \multicolumn{6}{l}{\textit{Standard Fact Verification (Manual \& Scraped)}} \\
    FEVER~\citep{thorne-etal-2018-fever} & 185K & 1 & Sentence & Crowdsourced annotation based on Wikipedia & High \\
    MultiFC~\citep{augenstein-etal-2019-multifc} & 35K & 1 & Document & Scraped from 26 fact-checking websites & High \\
    X-FACT~\citep{gupta-srikumar-2021-x} & 31K & 25 & Claim & Crowdsourced annotation of fact-checks & High \\
    AveriTeC~\citep{schlichtkrull2023averitec} & 4.5K & 1 & Claim & Expert human annotation with search queries & High \\
    FACTors~\citep{altuncu2025factors} & 118K & 1 & Claim & Scraped from IFCN \& Euro Code of Standards & High \\
    \midrule
    \rowcolor{gray!10} \multicolumn{6}{l}{\textit{LLM-Augmented \& Translated Verification}} \\
    MultiClaim~\citep{pikuliak-etal-2023-multilingual} & 206K & 39 & Claim & Aggregation + MT into target languages & Med \\
    FactLens~\citep{mitra-etal-2025-factlens} & 733 & 1 & Claim & LLM-based expansion + Human evaluation & Low \\
    MultiClaimNet~\citep{panchendrarajan-etal-2025-multiclaimnet} & 85K & 78 & Claim & Aggregation + LLM-based labeling & Med \\
    \midrule
    \rowcolor{gray!10} \multicolumn{6}{l}{\textit{KG-to-Text Alignment \& Textualization}} \\
    WebNLG~\citep{gardent-etal-2017-webnlg} & 45K & 2 & \textit{Synthetic} & Crowdsourcing + Machine Translation & High \\
    T-REx$^\dagger$~\citep{elsahar-etal-2018-rex} & 11M & 1 & Sentence & Distant Supervision (Wikidata--Wikipedia) & Med \\
    KELM~\citep{agarwal-etal-2021-knowledge} & 18M & 1 & \textit{Synthetic} & Seq2Seq generation (T5) from triples & Low \\
    \midrule
    \rowcolor{gray!10} \multicolumn{6}{l}{\textit{Knowledge Graph Benchmarks}} \\
    CoDEx~\citep{safavi-koutra-2020-codex} & 365K & -- & None & Wikidata triple extraction & High \\
    Wikidata5M~\citep{wang-etal-2021-kepler} & 5M & -- & Page & Wikidata + Wikipedia entity descriptions & Med \\
    OGB-WikiKG2~\citep{hu2021ogb} & 2.5M & -- & None & Extraction of triples (no text) & N/A \\
    Mintaka~\citep{sen-etal-2022-mintaka} & 20K & 8 & None & Human-authored complex questions over Wikidata & High \\
    KILT~\citep{petroni-etal-2021-kilt} & 44K & 1 & Passage & Wikipedia passage retrieval for multiple tasks & High \\
    Wikidata~\citep{vrandevcic2014wikidata} & $>$1B & $>$300 & None & Collaborative community curation & High \\
    \midrule
    \rowcolor{blue!10} \textbf{FactNet (Ours)} & \textbf{1.7B} & \textbf{316} & \textbf{Span/Pointer} & \textbf{Distant Supervision + Deterministic Alignment} & \textbf{Exact} \\
    \bottomrule
    \end{tabular}
    }
    % \vspace{-2.5mm}
\end{table*}

As Table~\ref{tab:comparison_main} illustrates, existing resources impose a three-way trade-off among \emph{scale}, \emph{grounding fidelity}, and \emph{linguistic coverage}\footnote{See Appendix~\ref{app:related_work_extended} for detailed analysis.}.
Knowledge bases such as Wikidata~\citep{vrandevcic2014wikidata} and Wikidata5M~\citep{wang-etal-2021-kepler} offer queryable structure at scale but provide no textual evidence, leaving RAG systems without the supervision signal needed to learn fact-to-text retrieval.
Grounded datasets such as FEVER~\citep{thorne-etal-2018-fever} and AveriTeC~\citep{schlichtkrull2023averitec} rely on manual curation, which limits their scale to thousands of instances and confines coverage largely to English.
Alignment resources such as T-REx~\citep{elsahar-etal-2018-rex} and KELM~\citep{agarwal-etal-2021-knowledge} bridge knowledge graphs and text, yet T-REx covers only English with sentence-level granularity, and KELM generates synthetic text rather than preserving authentic source documents.
Synthetic expansion via machine translation~\citep{chang-etal-2023-xfever,pikuliak-etal-2023-multilingual} or LLM generation~\citep{chung2025beyond,panchendrarajan-etal-2025-multiclaimnet} increases multilingual coverage but introduces translation artifacts and severs the connection to human-authored documents~\citep{koppel-ordan-2011-translationese,shumailov2023curse}.

One might ask why a Wikipedia-derived resource is useful when LLMs are trained on Wikipedia.
Recent evidence shows that LLMs hallucinate on well-established facts even within their training domain~\citep{ren-etal-2025-investigating,wang2025survey}, and factual knowledge degrades sharply for non-English languages~\citep{singhal-etal-2024-multilingual,fierro-etal-2025-multilingual}.
Moreover, \emph{knowing} a fact is not the same as being able to \emph{ground} it in retrievable evidence. This grounding capability is precisely what RAG systems must learn, and it requires resources where facts are explicitly linked to textual manifestations.

To bridge this gap, we introduce \textbf{FactNet}, a billion-scale multilingual graph that couples Wikidata statements with \emph{evidence pointers} derived from native Wikimedia dumps across 316 languages\footnote{We utilize Wikimedia dumps dated 2025-11-01.}.
FactNet employs a deterministic construction pipeline that ensures every evidence pointer is traceable to a specific location in the source data, without relying on stochastic inference or synthetic generation.
The graph is organized into three coupled layers (Figure~\ref{fig:structure}):
(1) \textbf{FactStatement:} An atomic, language-neutral unit representing a Wikidata claim, including qualifiers and references.
(2) \textbf{FactSense:} A grounded realization within a specific Wikipedia edition, linking to concrete textual evidence (sentence, infobox field, or table cell) with byte-level offsets.
(3) \textbf{FactSynset:} An equivalence class that groups FactStatements expressing the same underlying fact across languages and surface forms.

FactNet aggregates 1.7B FactStatements and 3.01B FactSenses into 1.55B FactSynsets across 316 languages, together with 3.69B rule-derived relational signals for downstream reasoning.
To demonstrate FactNet's utility as a benchmark, we construct \textbf{FactNet-Bench}, an evaluation suite for Knowledge Graph Completion (KGC), Multilingual KG Question Answering (MKQA), and Multilingual Fact Checking (MFC), with fixed splits, leakage controls, and baselines.

In summary, our contributions are as follows:
(i) \textbf{FactNet}, the first billion-scale resource that couples structured knowledge with native multilingual textual evidence;
(ii) a deterministic construction pipeline that avoids synthetic artifacts while achieving unprecedented scale; and
(iii) \textbf{FactNet-Bench}, a benchmark suite with systematic leakage controls for evaluating grounded factual reasoning.
\section{FactNet: Design and Construction}
\label{sec:factnet_construction}

\begin{figure*}[t]
    \centering
    \includegraphics[width=\textwidth]{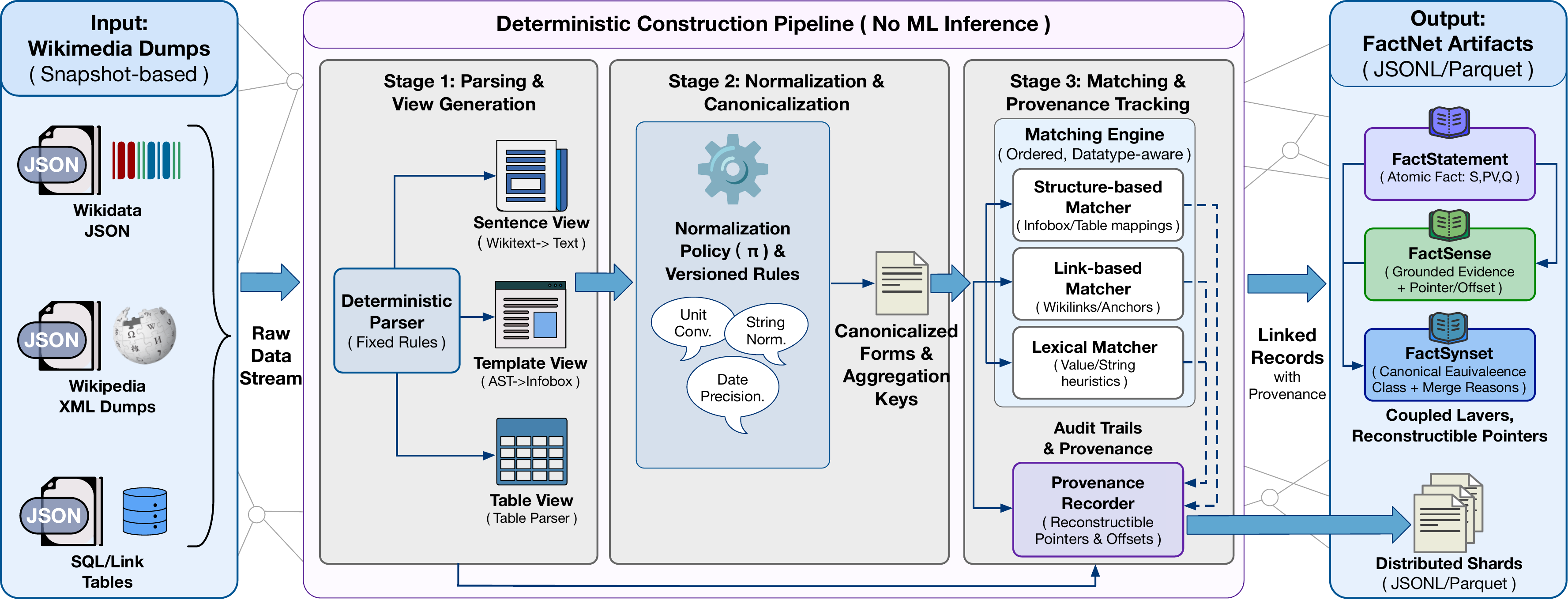}
    \caption{\textbf{FactNet construction workflow.} The pipeline processes dumps through three deterministic stages: (1) view extraction, (2) statement canonicalization, and (3) evidence matching. Every generated FactSynset and FactSense retains a stable provenance link to its source snapshot.}
    \label{fig:pipeline}
    \vspace{-1.5em}
\end{figure*}

FactNet aligns structured atomic assertions from Wikidata with grounded textual evidence from Wikipedia via provenance pointers.
Three principles govern the architecture:
(1) \emph{Cross-lingual Unification} via stable Wikidata identifiers;
(2) \emph{Policy-driven Canonicalization} for statement equivalence; and
(3) \emph{Deterministic Grounding}, where fact-to-text alignments rely on offsets that are reproducible from raw data dumps.
Figure~\ref{fig:pipeline} illustrates the construction workflow.

%%% figure

\textbf{Scope and Provenance.}
FactNet relies solely on Wikimedia dumps (Wikidata JSON, Wikipedia XML, and SQL link tables), forgoing stochastic inference or external curation.
All identifiers and offsets are defined relative to specific dump versions recorded in the build manifest (Appendix~\ref{app:reproducibility}).
Every record retains original source identifiers (\texttt{statement\_id}, \texttt{page\_id}, \texttt{revision\_id}).

\textbf{Language Coverage.}
For a target Wikipedia edition $\ell$, we ground statements whenever the subject entity contains an explicit sitelink to $\ell$.
A conservative fallback retrieves pages only if a deterministically normalized title resolves to exactly one non-disambiguation page (Appendix~\ref{app:factnet_factsense}).
Grounding is strictly scoped to the subject page; evidence from auxiliary pages is excluded.

\subsection{Data Model}
The FactNet schema distinguishes among atomic assertions, grounded evidence, and equivalence classes, serialized in canonical JSON with deterministic identifier hashing (Appendix~\ref{app:factnet_schema}).
\begin{itemize}[
    leftmargin=*,
    itemsep=2pt,
    topsep=-2pt,
    parsep=0pt,
    partopsep=0pt
]
    \item \textbf{FactStatement}: An atomic Wikidata claim indexed by \texttt{statement\_id}, comprising a subject-property-value tuple $(S, P, V)$, a multiset of qualifiers $Q$, references, and a \emph{rank} (\texttt{preferred}, \texttt{normal}, or \texttt{deprecated}).
    \item \textbf{FactSense}: A grounded mention within Wikipedia, storing language, provenance, evidence-unit type (\texttt{SENTENCE}, \texttt{INFOBOX\_FIELD}, or \texttt{TABLE\_CELL}), and an \texttt{evidence\_pointer} uniquely identifying the evidence span (Section~\ref{sec:offsets_main}).
    \item \textbf{FactSynset}: An equivalence class of FactStatements expressing the same fact across different surface forms or languages, recording member IDs and \texttt{merge\_reasons} for any relaxation beyond strict equivalence.
    \item \textbf{RelationEdge}: Typed, rule-derived connections between FactSynsets with structured provenance enabling re-derivation from released rule sets.
\end{itemize}

\subsection{Statement Canonicalization into FactSynsets}
\label{sec:canonicalization}

FactSynsets group statements that express the same fact with minor surface differences (e.g., time precision or unit conventions) or across language editions, enabling reasoning over \emph{facts} rather than \emph{surface forms}.

Let a statement be $f=(S,P,V,Q)$. FactSynsets are generated via a canonical \texttt{aggregation\_key}:
\begin{equation*}
\texttt{aggregation\_key} = S \concat P \concat \mathrm{NormValue}(P,V;\pi) \concat \mathrm{NormQuals}(Q;\pi)\, ,
\end{equation*}
where $\concat$ denotes string concatenation and $\mathrm{NormQuals}$ is order-invariant, deterministically sorting qualifiers by (PID, normalized value).
By default, FactNet merges only strictly equivalent normalized statements.
Semantic relaxations (time-precision truncation, unit conversion, and coordinate rounding) are applied only when authorized by policy $\pi$, and each relaxation is accompanied by a machine-readable \texttt{merge\_reason} (Appendix~\ref{app:factnet_normalization} and~\ref{app:factnet_synsets}).

\subsection{Grounding: FactSense Extraction}
\label{sec:offsets_main}
FactSenses align FactStatements to Wikipedia evidence using deterministic preprocessing and reconstructible pointers.

\textbf{Preprocessing Views.}
Three provenance-stable views are generated from raw wikitext:
(1) \emph{Sentence View}, via deterministic template stripping and segmentation;
(2) \emph{Template View}, via AST-based extraction\footnote{We employ \texttt{mwparserfromhell} (\url{https://github.com/earwig/mwparserfromhell}) for wikitext parsing.} of infobox parameters; and
(3) \emph{Table View}, via structural parsing of table cell content.
Full template expansion is deliberately avoided because recursive rendering introduces noise and offset instability.

\textbf{Evidence Pointers and Offsets.}
Each FactSense includes an \texttt{evidence\_pointer} anchored by (\texttt{page\_id}, \texttt{revision\_id}, view type, unit locator).
Unit locators correspond to structural indices (e.g., \texttt{sentence\_index}, \texttt{template\_path+param}, or \texttt{table\_id,row,col}).
Offsets are deterministic character offsets (Unicode codepoint indices) relative to the normalized evidence string, enabling exact re-localization from source dumps (Appendix~\ref{app:factnet_offsets}).
This byte-level precision distinguishes FactNet from prior distant-supervision resources such as T-REx, which provide only sentence-level alignment.

\textbf{Multilingual Segmentation.}
Stanza~\citep{qi-etal-2020-stanza} is used for 70 languages; for the remaining 246, a deterministic rule-based segmenter that relies on punctuation and Unicode boundaries is employed.
Both backends yield comparable grounding precision (0.926 vs.\ 0.914; Section~\ref{sec:stats_quality}), and all segmentation choices are versioned within each FactSense record (Appendix~\ref{app:factnet_languagepacks}).

\textbf{Alignment Strategy.}
We employ distant supervision~\citep{mintz-etal-2009-distant,elsahar-etal-2018-rex} with ordered, datatype-aware matchers:
(1) \emph{Structure-based matching}~\citep{auer2007dbpedia,suchanek2007yago}, which utilizes infobox and table parameter mappings for literal values;
(2) \emph{Link-based matching}, which aligns entity-valued $V$ (Wikilinks or anchors resolving to the QID of $V$); and
(3) \emph{Lexical matching}, which targets literal values (time, quantity, coordinates, strings) within sentence or table contexts.
Our alignment verifies the presence of known Wikidata values rather than performing full relation extraction from text.
Evidence units are deduplicated by prioritizing the highest-confidence \texttt{match\_type}, which ranges from \texttt{WIKILINK\_ENTITY} (highest) through \texttt{INFOBOX\_FIELD} and \texttt{LEXICAL\_VALUE} to \texttt{LEAD\_WEAK} (lowest), while preserving alternative hits as metadata.

\subsection{Relational Structure}
\textbf{RelationEdges} provide structural connectivity with strict derivability:
(1) \emph{Direct Joins}, which connect synsets where the object entity of one is the subject of another, filtered by a descriptive-property allowlist;
(2) \emph{Schema-based Relations}, which are induced by a released \texttt{PROPERTY\_RELATION\_MAP} with bounded traversal (max hop $=$ 2); and
(3) \emph{Conflict Signals}, namely \texttt{POTENTIAL\_CONFLICT} edges derived from logical constraints (functional property violations, incompatible temporal intervals), modeled as signals rather than asserted contradictions.
All mapping files are versioned and categorized by reliability tiers (Appendix~\ref{app:factnet_relations}).

\subsection{Format, Release, and Licensing}
FactNet is fully deterministic given fixed input dumps, parser versions, and configuration.
We release all components necessary for independent reconstruction, including schemas, normalization policy $\pi$, language packs, mapping resources, and build manifests (Appendix~\ref{app:reproducibility}).
The dataset is distributed in sharded, compressed formats (e.g., Parquet) along with indexing scripts (Appendix~\ref{app:release_format}).
A curated subset covering the top 50 languages with strong-evidence synsets reduces the on-disk footprint from 894 GB to approximately 120 GB.

\textbf{Licensing.} Wikidata content is released under CC0; Wikipedia textual content under CC BY-SA.
The default distribution contains structural IDs, pointers, and offsets; raw evidence text is provided in a separate pack under CC BY-SA with mandatory attribution (Appendix~\ref{app:licensing}).
\section{Resource Statistics and Quality Assessment}
\label{sec:stats_quality}

We evaluate FactNet along three axes:
(i) scale and distributional properties,
(ii) grounding fidelity, comprising provenance stability, precision, and recall, and
(iii) structural integrity of canonicalization and edge derivation.
All statistics are derived from the \texttt{2025-11-01} Wikimedia snapshots using the default build configuration.

\subsection{Scale, Definitions, and Long-Tail Structure}
Table~\ref{tab:at_a_glance} summarizes aggregate scale.
The corpus consolidates 1.7B FactStatements (12.1K properties) into 1.55B FactSynsets, supported by 3.01B FactSenses across 316 languages and 3.69B rule-derived RelationEdges.

%%% Overall statitics table
\begin{wraptable}[13]{r}{0.5\linewidth}
    \vspace{-2em}
    \centering
    \caption{\textbf{FactNet statistics (\texttt{2025-11-01}).} Scale, coverage, and provenance.}
    \label{tab:at_a_glance}
    \resizebox{\linewidth}{!}{
    \begin{tabular}{lr}
    \toprule
    \textbf{Metric} & \textbf{Value} \\
    \midrule
    FactStatements / Properties & 1.70\,B / 12.1\,K \\
    FactSynsets & 1.55\,B \\
    FactSenses / RelationEdges & 3.01\,B / 3.69\,B \\
    \midrule
    Evidence-bearing synsets & 1.05\,B \; (67.93\%) \\
    Strong-evidence synsets & 0.81\,B \; (52.48\%) \\
    Multilingual synsets (evidence; $\ge2$ langs) & 0.49\,B \; (31.84\%) \\
    Multilingual synsets (sitelink; $\ge2$ langs) & 0.95\,B \; (61.19\%) \\
    \midrule
    Statements with $\ge$1 reference & 72.27\% \\
    Statements with $\ge$1 qualifier & 36.04\% \\
    \midrule
    On-disk footprint (Parquet) & 894\,GB \\
    Provenance re-localization (1M sample) & 99.63\% exact \\
    \bottomrule
    \end{tabular}}
    \vspace{-1.5em}
\end{wraptable}

\textbf{Evidence Strata.}
We stratify synsets into four categories:
(1) \emph{Evidence-bearing}, containing at least one FactSense;
(2) \emph{Strong-evidence}, supported by high-precision extraction (\texttt{WIKILINK\_ENTITY} or \texttt{INFOBOX\_FIELD});
(3) \emph{Multilingual (Evidence)}, with extracted FactSenses in $\ge$2 languages; and
(4) \emph{Multilingual (Sitelink)}, where the subject entity possesses sitelinks to $\ge$2 Wikipedia editions.

\textbf{Distributional Skew.}
The supervision reflects Wikipedia's heavy-tailed distribution~\citep{kaffee2017glimpse}: the top five languages contribute 63.4\% of FactSenses (76.1\% for the top ten), while the bottom 200 account for only 2.7\%.
The effective language count ($N_{\text{eff}} = 2^{H}$, where $H$ is Shannon entropy) is approximately 18.7 to 20.4, far exceeding any existing resource but substantially less than the raw 316 count suggests.
Property coverage exhibits similar sparsity: 31.8\% of properties have at least 100 evidence-bearing synsets, but only 9.4\% exceed 10{,}000 (Appendix~\ref{app:summary_full}).
Our evaluation employs stratified sampling across language tiers, scripts, and match types.

\textbf{The Evidence Gap.}
While 61.19\% of synsets are multilingual via sitelink connectivity, only 31.84\% possess extracted evidence in multiple languages (Table~\ref{tab:at_a_glance}).
A funnel analysis reveals that for high-resource languages, the yield rate (at least one FactSense given a retrievable subject page) is 0.79, compared to 0.36 for low-resource languages.
The primary bottleneck lies in within-page matching rather than retrieval failures (Appendix~\ref{app:funnel}).

\subsection{Content Distribution and Representation}
FactNet inherits the topical and societal biases of its source knowledge bases.
We provide diagnostic metrics to support responsible benchmarking:
(1) Topical distribution skews toward humans, geographic entities, and organizations (58\% of evidence-bearing synsets).
(2) Demographic imbalance~\citep{zhang2021quantifying} among subjects typed as \emph{human} (Q5) with \emph{sex\_or\_gender} (P21) reveals a distribution of 77\% male, 22\% female, and 1\% other or unknown.
(3) Geographic concentration~\citep{das2023diversity} for coordinate-bearing subjects favors Europe and North America (52\%), followed by East Asia (17\%) and South Asia (8\%).
Detailed statistics by language tier are in Appendix~\ref{app:representation_bias}.

\subsection{Audit Protocol and Evaluation Estimands}
Because automated heuristics cannot reliably verify semantic entailment, we conducted a human-in-the-loop audit (Appendix~\ref{app:audit_protocol}).
Our primary estimand is \emph{corpus-level FactSense precision}: the frequency-weighted probability that a randomly selected FactSense is semantically correct.
We employed stratified cluster sampling at the $(\ell, \texttt{page\_id})$ level with design weights.
Inter-annotator agreement was robust (Krippendorff's $\alpha=0.82$ for correctness; range $\in [0.74, 0.90]$\footnote{Values $\ge 0.80$ indicate high reliability.}).
Among adjudicated items (9.8\% abstain rate), machine translation\footnote{NLLB-200~\citep{costa2022no}; annotators used MT only as an assistive signal.} was consulted in 41\% of cases, yielding a negligible precision difference (no more than 0.6 percentage points).

%%%% maual audit table
\begin{wraptable}[8]{r}{0.5\linewidth}
    \vspace{-2em}
    \centering
    \caption{\textbf{FactSense grounding precision.} Manual audit by \texttt{match\_type} with design-weighted overall (95\% Wilson CI).}
    \label{tab:grounding_precision_compact}
    \resizebox{\linewidth}{!}{
    \begin{tabular}{l r r r}
    \toprule
    \textbf{Match type} & \textbf{Share} & \textbf{Precision} & \textbf{95\% CI} \\
    \midrule
    \texttt{WIKILINK\_ENTITY} & 35.0\% & 0.973 & [0.964, 0.980] \\
    \texttt{INFOBOX\_FIELD} & 20.0\% & 0.944 & [0.932, 0.955] \\
    \texttt{LEXICAL\_VALUE} & 35.0\% & 0.889 & [0.873, 0.904] \\
    \texttt{LEAD\_WEAK} & 10.0\% & 0.808 & [0.778, 0.836] \\
    \midrule
    \rowcolor{blue!10}
    \textbf{Overall (design-weighted)} & 100\% & \textbf{0.921} & \textbf{[0.913, 0.929]} \\
    \bottomrule
    \end{tabular}}
\end{wraptable}

\subsection{Grounding Quality of FactSenses}
\textbf{Provenance Stability.}
Across a stratified sample of one million items, exact re-localization achieved 99.63\% (Table~\ref{tab:at_a_glance}).
Both segmentation backends demonstrated high stability (Stanza: 99.71\%, rule-based: 99.54\%) and comparable precision (0.926 vs.\ 0.914), confirming deterministic segmentation viability for low-resource languages (Appendix~\ref{app:integrity}).

\textbf{Grounding Precision.}
Across $n=4{,}200$ items spanning 316 languages, the design-weighted precision is 0.921 (95\% CI [0.913, 0.929]).
\texttt{WIKILINK\_ENTITY} and \texttt{INFOBOX\_FIELD} matchers account for 55\% of senses and achieve precision exceeding 0.94 (Table~\ref{tab:grounding_precision_compact}), which motivates our \emph{strong-evidence} filter.
Precision in Tier-3 languages (0.883) is lower than in Tier-1 (0.938), a gap that reflects template and link sparsity.

\textbf{Missingness and Recall Lower Bounds.}
The estimated false-negative rate is 24\% (95\% CI [20\%, 28\%]; Appendix~\ref{app:recall_lower_bound}).
Missingness is systematic: temporal and relational properties (e.g., \texttt{P569} and \texttt{P26}) are disproportionately affected (32\% ungrounded vs.\ 19\% for quantity properties), primarily owing to paraphrastic phrasing.
The rate increases from 17\% (Tier 1) to 36\% (Tier 3), consistent with the funnel analysis.
We provide \texttt{ungrounded\_reason} codes for all unmapped facts (Appendix~\ref{app:grounding_missingness}).

\textbf{Fine-Grained Failure Analysis.}
Over 300 sampled ungrounded facts, temporal properties exhibit the highest ungrounded rate (38\%), followed by relational (34\%), quantity (19\%), geographic (16\%), and identifier properties (11\%).
The interaction between property type and language tier amplifies these disparities: temporal properties in Tier-3 languages reach 51\% ungrounded compared to 22\% in Tier-1.
The primary failure causes are paraphrastic phrasing (45\%), template-mapping gaps (28\%), missing entity links (18\%), and other causes (9\%).
Template-mapping gaps concentrate disproportionately in Tier-2 and Tier-3 languages (41\% of Tier-3 failures compared to 19\% in Tier-1), whereas paraphrastic phrasing dominates in Tier-1 (52\%).
We release the annotated sample with property-type and tier labels (Appendix~\ref{app:false_negative_detail}).

\subsection{Canonicalization and Structural Quality}
\textbf{Synset Integrity.}
Based on $n=1{,}000$ synsets per category, false-merge rates are low: 0.005 [0.002, 0.011] for strict merges and 0.017 [0.011, 0.026] for policy-relaxed merges.
Policy-relaxed merges account for only 1.3\% of the total; the most common relaxations are time-precision truncation (41\%) and unit conversion (23\%).
Policy-relaxed merges affect only 1.3\% of the total; the most common relaxations are time-precision truncation (41\%) and unit conversion (23\%).

\textbf{RelationEdges: Precision vs.\ Depth.}
Manual audits ($n=3{,}000$) reveal that precision declines with derivation depth: \textbf{0.953} for direct joins, \textbf{0.918} for 1-hop, and \textbf{0.882} for 2-hop relations.
We recommend using the provided filters (e.g., hop cap $=$ 1) for noise-sensitive tasks.
The \texttt{POTENTIAL\_CONFLICT} signal (2.69\% of synsets) achieves 0.742 precision and serves as a triage mechanism (Appendix~\ref{app:edges}).

\textbf{Pipeline Integrity.}
FactNet is fully deterministic given fixed input dumps, parser versions, and configuration~\citep{gebru2021datasheets}.
We provide complete schemas, policy definitions, and mapping resources for independent reconstruction; integrity violations are logged explicitly (Appendix~\ref{app:reproducibility}).
We provide complete schemas, policy definitions, and mapping resources for independent reconstruction; integrity violations are logged explicitly (Appendix~\ref{app:reproducibility}).
\section{FactNet-Bench: Tasks and Experiments}
\label{sec:benchmarks_experiments}

We introduce FactNet-Bench, a benchmark suite that evaluates three capabilities:
(i) Knowledge Graph Completion (KGC;~\citealp{bordes2013translating}), which assesses reasoning over canonicalized facts;
(ii) Multilingual KG Question Answering (MKQA;~\citealp{longpre-etal-2021-entity}), which evaluates the generation of executable logical forms grounded in FactNet identifiers; and
(iii) Multilingual Fact Checking (MFC;~\citealp{thorne-etal-2018-fever}), which tests veracity prediction using FactSense-grounded evidence.
Task statistics appear in Appendix~\ref{app:bench_statistic}.

\subsection{Benchmark Design}
\label{sec:bench_design}

All instances, splits, and evaluation artifacts are generated from a frozen FactNet snapshot, eliminating any dependency on external endpoints.
We release the complete construction pipeline, split definitions, and scoring scripts to enable exact replication (Appendix~\ref{app:reproducibility}).
Splits are generated via stable hashing of unique \texttt{synset\_id}s, ensuring that all facts related to a specific synset reside within a single partition. Task-specific projections undergo de-duplication to eliminate multi-edge artifacts (Appendix~\ref{app:kgc_graph}).
Any \textsc{FactSense} aligned to dev/test synsets is excluded from training-time retrieval pools and entity descriptions.
For KGC, query-time predicate masking prevents trivial completion by extraction (Appendix~\ref{app:text_masking}).
Rule-derived \textsc{RelationEdge}s used by GNN-based systems must be constructed solely from the training split, maintaining a strict inductive setting (Appendix~\ref{app:relationedge_use}).

%%% Experiments on FactNet-Bench Figure
\begin{figure*}[t]
    \centering
    \includegraphics[width=\linewidth]{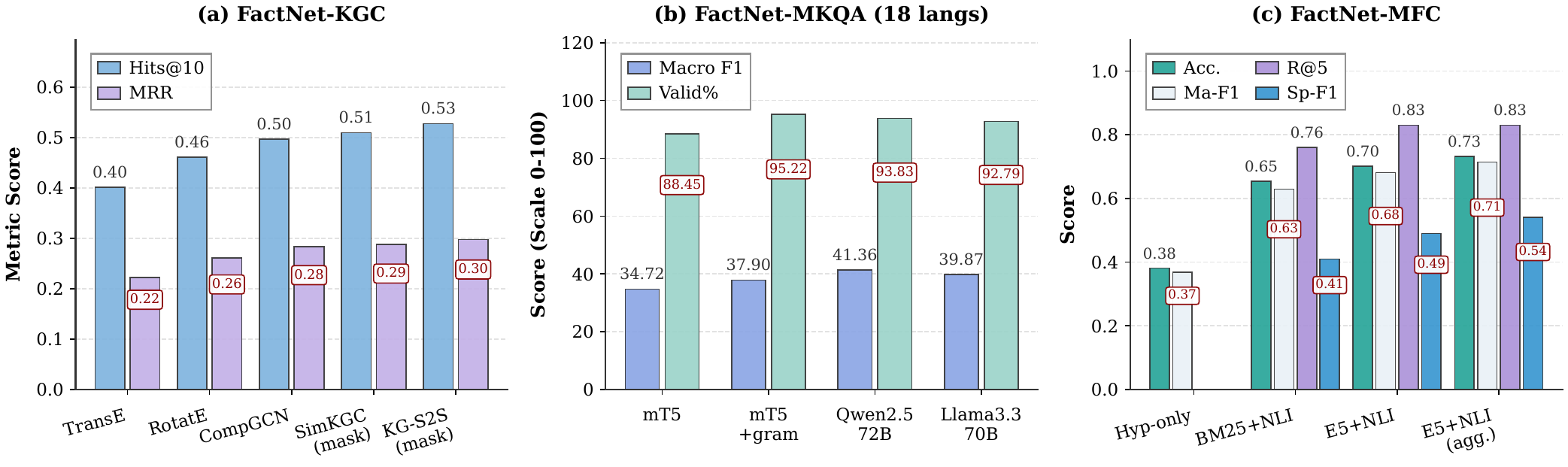}
    \caption{Results on FactNet-Bench: (a) KGC under leakage control, (b) MKQA semantic parsing (18 languages), (c) MFC verification and evidence quality. Error bars indicate std.\ over 3 seeds.}
    \label{fig:res_main}
    \vspace{-1em}
\end{figure*}

\subsection{Tasks and Evaluation Metrics}
\label{sec:eval_protocols}

We report results as mean $\pm$ std over three random seeds.

\textbf{KGC (Entity Link Prediction).}
Filtered, fully-ranked link prediction on an entity-centric graph induced from synsets with entity-valued main arguments~\citep{liu-etal-2025-enhancing-large,luo-etal-2025-gltw}.
Metrics: MRR~\citep{craswell2016mean} and Hits@10.
Qualifiers are not predicted, thereby isolating structural reasoning (Appendix~\ref{app:kgc_graph}).

\textbf{MKQA (Multilingual Executable Semantic Parsing).}
Instances pair natural-language questions with restricted executable logical forms, covering 1-hop and constrained 2-hop queries; invalid parses receive a score of zero.
Metric: Macro F1 between predicted and gold answers after normalization~\citep{tian-etal-2025-compkbqa} (Appendix~\ref{app:mkqa_eval}).

\textbf{MFC (Closed-Context Fact Checking).}
Given a claim, systems predict veracity labels (\textsc{Supported}, \textsc{Refuted}, \textsc{NEI}) on the basis of evidence in the frozen snapshot, retrieving \textsc{FactSense} units and identifying character-offset spans within them.
Metrics: label Accuracy, Macro F1, evidence Recall@5, and span-level Evidence F1~\citep{thorne-etal-2018-fever} (Appendix~\ref{app:mfc_contract}).

\subsection{Baselines}
\label{sec:baselines}

\textbf{KGC.}
Structural approaches (TransE;~\citealp{bordes2013translating}, RotatE;~\citealp{sun2019rotate}, CompGCN;~\citealp{vashishth2019composition}), text-aware architectures (SimKGC;~\citealp{wang-etal-2022-simkgc}, KG-S2S;~\citealp{chen-etal-2022-knowledge}), and LLM-integrated methods (GLTW;~\citealp{luo-etal-2025-gltw}, SAT;~\citealp{liu-etal-2025-enhancing-large}).
Text-aware models employ leakage-controlled entity descriptions with predicate masking (Appendix~\ref{app:text_masking}).

\textbf{MKQA.}
Fine-tuned mT5 parsers with and without grammar-guided decoding~\citep{srivastava2024mst5}, alongside open-weight LLMs (Qwen-2.5-72B;~\citealp{yang2025qwen3}, LLaMA-3.3-70B;~\citealp{grattafiori2024llama}) using 5-shot prompts with constrained decoding (Appendix~\ref{app:mkqa_prompt}).

\textbf{MFC.}
A hypothesis-only baseline together with full pipelines that pair retrieval (BM25;~\citealp{robertson2009probabilistic}, E5-large;~\citealp{wang2024multilingual}, translation-assisted) with an XLM-R NLI verifier~\citep{conneau-etal-2020-unsupervised} (Appendix~\ref{app:mfc_contract}).

\subsection{Results and Analysis}
\label{sec:results}

\textbf{(I) KGC.}
Figure~\ref{fig:res_main}(a) presents performance across model categories.
Structural baselines exhibit the expected hierarchy (TransE $<$ RotatE $<$ GNN), confirming benchmark validity.
Text-aware methods yield further gains (KG-S2S improves upon CompGCN by 0.014 MRR); LLM-integrated methods achieve the highest performance (GLTW: 0.386 MRR).
A diagnostic ablation confirms the necessity of leakage control: without masking, KG-S2S MRR increases anomalously from 0.298 to 0.351, indicating that the task degenerates into information extraction.
Train-only \textsc{RelationEdge}s enhance GNN performance (Appendix~\ref{app:relationedge_use}).

\textbf{(II) MKQA.}
Figure~\ref{fig:res_main}(b) identifies executability as the primary bottleneck.
Grammar-guided decoding improves Macro F1 by 3.2 points and raises validity from 88.5\% to 95.2\%.
Prompted LLMs achieve the highest semantic accuracy (Qwen-2.5: 41.4 Macro F1), yet grammar-guided fine-tuned models maintain a slight edge in interface compliance (95.2\% vs.\ 93.8\% validity).
Performance disparities narrow in low-resource languages (Appendix~\ref{app:mkqa_langs}).

\textbf{(III) MFC.}
The hypothesis-only diagnostic achieves 0.381 accuracy; evidence-based systems surpass this by 0.27 to 0.35 points (Figure~\ref{fig:res_main}(c)).
Dense retrieval (E5-large) substantially benefits verification accuracy (0.701 vs.\ 0.654) and evidence quality (R@5: 0.83 vs.\ 0.76; Span F1: 0.49 vs.\ 0.41).
Top-5 aggregation further improves accuracy and span F1.

\subsection{Multilingual Analysis}
\label{subsec:multilingual_analysis}

\textbf{KGC by Language Tier.}
Structural methods show stable performance across tiers, whereas text-aware methods exhibit larger variance (SimKGC: 0.312 MRR on Tier-1 $\to$ 0.271 on Tier-3).
GLTW exhibits the smallest cross-tier gap (0.386 to 0.359), suggesting that combining structural encoding with LLM reasoning mitigates text sparsity.

\textbf{Cross-lingual Transfer.}
Training SimKGC on only English entity descriptions and evaluating on all tiers retains 78\% of English performance on Tier-2 and 64\% on Tier-3, demonstrating that the shared FactSynset structure enables meaningful transfer without target-language text.

\subsection{External Validation}
\label{subsec:external_validation}

\textbf{(IV) FactNet-Augmented RAG.}
A RAG system (E5-large retriever + FiD reader) trained on FactNet \textsc{FactSense} evidence outperforms a Wikipedia-dump baseline on TriviaQA~\citep{joshi-etal-2017-triviaqa} (71.2 vs.\ 67.8 EM, +3.4) and on FEVER~\citep{thorne-etal-2018-fever} (accuracy 87.9 vs.\ 85.7, +2.2; evidence F1 76.8 vs.\ 72.3, +4.5).
For cross-lingual queries, FactNet improves evidence F1 by 7.5 points (Appendix~\ref{app:rag_detail}).

\textbf{(V) Cross-lingual Zero-shot Verification.}
Training on Tier-1 claims and evaluating zero-shot on X-FACT~\citep{gupta-srikumar-2021-x} and MultiClaim~\citep{pikuliak-etal-2023-multilingual}, FactNet-trained models outperform MT-projection (NLLB-200~\citep{costa2022no}) by 7.5 and 7.3 Macro F1 points, respectively.
The advantage is largest for languages with poor translation quality, where MT-projection introduces label noise while FactNet's native multilingual evidence avoids these distortions (Appendix~\ref{app:xling_detail}).

\subsection{LLM Hallucination Probing}
\label{subsec:hallucination_probing}

We sample 2,000 \textsc{FactSynset} facts from the test split, stratified by property type and language tier, and prompt GPT-4o~\citep{hurst2024gpt} and Qwen-2.5-72B~\citep{yang2025qwen3} with natural-language questions.
Responses are classified as \emph{hallucinated} (factually incorrect), \emph{correct} (matches $V$), or \emph{refused}.

\textbf{(VI) Results.}
GPT-4o hallucination rates: 22.4\% (English), 35.7\% (Tier-2), 44.1\% (Tier-3).
Qwen-2.5-72B: 27.8\%, 39.2\%, 48.3\%.
Both models refuse fewer than 3\% of queries.
Temporal facts are most error-prone (34.2\% for GPT-4o on English), followed by relational (31.7\%); geographic (12.8\%) and quantity (16.5\%) facts prove more reliable.
These results challenge the assumption that LLMs encode reliable factual knowledge from Wikipedia (Appendix~\ref{app:hallucination_detail}).

\subsection{Evidence Type Ablation}
\label{subsec:evidence_ablation}

We compare full-evidence against \texttt{SENTENCE}-only configurations (excluding infobox and table data).

\textbf{(VII) KGC.}
Removing infobox and table evidence reduces entity description coverage from 16.2 to 9.4 units per entity.
Text-aware models decline: GLTW drops by 0.039 MRR (0.386 to 0.347), SimKGC by 0.031 (0.312 to 0.281); structural models remain unaffected.

\textbf{(VIII) MFC.}
With E5-large retrieval, label accuracy drops by 0.039 (0.701 to 0.662), evidence R@5 drops by 0.12 (0.83 to 0.71), and Span F1 drops by 0.06 (0.49 to 0.43).
Approximately 38\% of \textsc{Supported} claims have at least one gold \texttt{INFOBOX\_FIELD} evidence unit.
Sentence-only evidence reduces MFC evidence R@5 by 14.5\% relative to the full configuration (Appendix~\ref{app:evidence_ablation_detail}).
\section{Discussion and Future Works}
\label{sec:discussion_futurework}

\textbf{Addressing Circularity Concerns.}
A legitimate concern is that some Wikipedia content, particularly infoboxes, is populated from Wikidata, creating circular provenance.
Three factors mitigate this concern: (1) sentence-level evidence, which constitutes the majority of FactSenses, is human-authored; (2) FactNet explicitly distinguishes evidence types (\texttt{INFOBOX\_FIELD} vs.\ \texttt{SENTENCE}), allowing researchers to filter for human-authored evidence when circularity is a concern; and (3) the infobox-to-Wikidata circularity provides a high-precision alignment signal (0.944 precision for \texttt{INFOBOX\_FIELD}) that validates the pipeline.
For applications requiring provenance independence, we recommend the \texttt{SENTENCE}-only subset.

\textbf{Design Trade-offs.}
FactNet prioritizes precision over recall, guaranteeing reliability for benchmarking but limiting recall, particularly in long-tail languages with inconsistent structural templates.
The 24\% false-negative rate is not uniform: temporal and relational properties are disproportionately affected, introducing a systematic bias that users should account for.
FactNet also reflects the demographic and topical biases present in its source Wikimedia dumps (Appendix~\ref{app:discussion_limitations}).

\textbf{Future Directions.}
Future iterations will aim to bridge the coverage gap through controlled neuro-symbolic alignment strategies that enhance recall without sacrificing traceability, expand the schema to support complex n-ary relations, and implement diff-based update mechanisms to synchronize with the evolving knowledge landscape (Appendix~\ref{app:future_roadmap}).
The external validation experiments (\S\ref{subsec:external_validation}, \S\ref{subsec:hallucination_probing}) confirm that FactNet transfers effectively to external benchmarks and that LLMs still exhibit substantial hallucination rates on FactNet-sourced facts.
\section{Conclusion}
\label{sec:conclusion}

We present \textbf{FactNet}, a billion-scale multilingual resource that unifies structured Wikidata assertions with span-level textual evidence from Wikipedia across 316 languages.
By grounding every fact in authentic, human-authored content via deterministic construction, FactNet provides reliable supervision for RAG systems, fact verification, and multilingual reasoning without relying on synthetic generation.
Complementing the resource, \textbf{FactNet-Bench} establishes rigorous evaluation standards that penalize information leakage, reward provenance quality, and support cross-lingual analysis.
Experiments demonstrate that FactNet-Bench effectively differentiates structural, text-aware, and LLM-integrated approaches, while the cross-lingual FactSynset structure facilitates knowledge transfer across language tiers.
We publicly release FactNet, its construction pipeline, and the benchmark suite to advance research in multilingual grounded reasoning and trustworthy AI.

\small
\bibliographystyle{unsrt}
\bibliography{neurips_2026}

%%%%%%%%%%%%%%%%%%%%%%%%%%%%%%%%%%%%%%%%%%%%%%%%%%%%%%%%%%%%%%%%%%%%%%%%%%%%%%%
%%%%%%%%%%%%%%%%%%%%%%%%%%%%%%%%%%%%%%%%%%%%%%%%%%%%%%%%%%%%%%%%%%%%%%%%%%%%%%%
% APPENDIX
%%%%%%%%%%%%%%%%%%%%%%%%%%%%%%%%%%%%%%%%%%%%%%%%%%%%%%%%%%%%%%%%%%%%%%%%%%%%%%%
%%%%%%%%%%%%%%%%%%%%%%%%%%%%%%%%%%%%%%%%%%%%%%%%%%%%%%%%%%%%%%%%%%%%%%%%%%%%%%%
\clearpage
\newpage
\appendix
\onecolumn

%% 附录目录页
% 打印附录目录
\startcontents[appendices]
\printcontents[appendices]{l}{1}{%
    \section*{Appendix Contents}%
    \setcounter{tocdepth}{2}%
}

\newpage

\section{Extended Review of Related Work and Resource Analysis}
\label{app:related_work_extended}

In this section, we present a comprehensive analysis of the existing factual resource landscape summarized in Table~\ref{tab:comparison_main}. We examine the inherent trade-offs currently imposed on the research community, with particular attention to the tension among scale, multilingual coverage, and evidence provenance. We categorize prior work into four primary paradigms and delineate their respective limitations with respect to the construction of grounded generation systems.

\subsection{Human-Curated and Fact-Checking Resources}

The foundational paradigm for fact verification has traditionally relied on manual annotation or controlled data collection. FEVER\footnote{\url{https://fever.ai}}~\citep{thorne-etal-2018-fever} established a seminal schema for this task by associating claims with evidence sentences and verification labels. While FEVER provides high-quality, granular grounding, its construction required crowd-workers to draft claims manually based on Wikipedia introductions. This dependence on manual labor creates a significant bottleneck for scalability and impedes expansion into low-resource languages.

Subsequent initiatives have attempted to address the complexity limitations of early datasets. Resources such as AveriTeC~\citep{schlichtkrull2023averitec} and EX-FEVER~\citep{ma-etal-2024-ex} introduce multi-hop reasoning and real-world search scenarios. However, the requirement for expert annotation in these datasets restricts their volume to the range of thousands, rendering them insufficient for pre-training large-scale retrieval models. In a parallel vein, datasets derived from professional fact-checking portals~\citep{wadden-etal-2020-fact}, including MultiFC~\citep{augenstein-etal-2019-multifc} and X-FACT~\citep{gupta-srikumar-2021-x}, capture naturally occurring misinformation. Nevertheless, these resources often lack granular evidence pointers: they typically provide document-level evidence rather than precise sentence-level justification, and their topical coverage is frequently limited to transient news cycles rather than the encyclopedic breadth required for general LLM grounding.

\subsection{Synthetic Expansion via Translation and Generative Models}

To mitigate the scalability constraints of manual annotation, recent methodologies have adopted synthetic expansion strategies, primarily utilizing machine translation or large language models. Translation-based approaches, such as XFEVER~\citep{chang-etal-2023-xfever} and MultiClaim~\citep{pikuliak-etal-2023-multilingual}, employ projection techniques to extend English datasets into other languages. Although this strategy effectively increases linguistic coverage, it introduces two methodological concerns. First, it often results in translationese~\citep{koppel-ordan-2011-translationese}, where the linguistic patterns reflect the syntax of the source language rather than the fluency of the target. Second, it risks cultural misalignment, as claims pertinent to English-speaking contexts may lack relevance or supporting evidence in the local Wikipedia editions of the target languages.

More recently, studies such as MultiSynFact~\citep{chung2025beyond} and FactLens~\citep{mitra-etal-2025-factlens} have leveraged LLMs to generate both claims and evidence synthetically. While this approach achieves substantial scale, it fundamentally alters the nature of provenance: synthetic datasets prioritize internal consistency over factual alignment with external reality. Consequently, utilizing LLM-generated data to train verification systems may induce a circular dependency~\citep{shumailov2023curse}, whereby the verifier learns the parametric patterns of the generator model rather than acquiring the capability to ground claims in human-authored sources. In contrast, FactNet maintains strict adherence to authentic provenance, ensuring that every evidence span is derived directly from human-authored Wikipedia content.

\subsection{Knowledge Graph Alignments and Textualization}

A distinct line of research employs Distant Supervision (DS;~\citealp{mintz-etal-2009-distant}) to align knowledge graphs with textual corpora. T-REx~\citep{elsahar-etal-2018-rex} represents a prominent effort in this domain, aligning Wikidata triples to Wikipedia sentences via heuristic matching. However, T-REx is restricted to English, and its alignment algorithms occasionally yield false positives in sentences containing multiple entities. Furthermore, T-REx provides sentence-level alignment without byte-level offset reproducibility, making it impossible to relocate exact evidence spans from source dumps\footnote{T-REx relies on sentence segmentation and tokenization from older versions of NLP libraries. Re-processing the corpus with modern tools often shifts token indices, breaking the alignment map provided in the dataset.}.

KELM~\citep{agarwal-etal-2021-knowledge} adopts a generative paradigm by converting KG subgraphs into natural-language sentences. While valuable for data augmentation, KELM constitutes a corpus of synthetic text that does not reference actual occurrences of facts on the web, limiting its utility for training retrieval-augmented generation (RAG) systems that must navigate noisy, real-world documents.

\subsection{Knowledge Graph Benchmarks}

Several KG benchmarks have been constructed from Wikidata and Wikipedia for specific tasks. CoDEx~\citep{safavi-koutra-2020-codex} provides a compact benchmark for link prediction and recommendation over Wikidata triples but lacks textual evidence entirely. Wikidata5M~\citep{wang-etal-2021-kepler} integrates Wikidata with Wikipedia entity descriptions to enable text-enhanced link prediction, but provides only page-level descriptions rather than span-level evidence and covers only English. Mintaka~\citep{sen-etal-2022-mintaka} offers complex, multilingual questions over Wikidata but does not provide textual grounding. KILT~\citep{petroni-etal-2021-kilt} prioritizes provenance by linking tasks to specific Wikipedia passages, supporting diverse tasks including slot filling, QA, and fact checking, but is limited to English and contains only 44K instances. OGB-WikiKG2~\citep{hu2021ogb} provides large-scale triple extraction for benchmarking graph neural networks but includes no textual content. Finally, massive KGs such as DBPedia~\citep{auer2007dbpedia}, Freebase~\citep{bollacker2008freebase}, and Wikidata~\citep{vrandevcic2014wikidata} provide the requisite scale and structure but lack textual grounding: a structural tuple encodes a fact but offers no linguistic signal regarding how that fact is expressed in natural language across diverse contexts.

\subsection{FactNet in the Resource Landscape}

The analysis above reveals that existing resources occupy distinct positions in a three-dimensional space defined by \emph{scale}, \emph{grounding fidelity}, and \emph{linguistic coverage}, and that no prior resource combines strength in all three dimensions. Manual datasets such as FEVER offer high grounding fidelity but low scale and monolingual coverage. Synthetic datasets such as KELM offer scale but compromised authenticity. KG benchmarks such as CoDEx and Wikidata5M offer structure but lack textual grounding. KILT offers provenance but at limited scale and only in English.

FactNet occupies a previously vacant position in this space by applying the scale of distant supervision within a reproducible, provenance-preserving pipeline across 316 languages. By treating Wikipedia as a structured XML tree for indexing\footnote{We refer to the raw XML dumps provided by the Wikimedia Foundation, available at \url{https://dumps.wikimedia.org}.} rather than a corpus for scraping, we achieve the magnitude of knowledge graphs while preserving the grounding granularity of verification datasets and the reproducibility that synthetic approaches lack.
\section{Implementation Details for FactNet Construction and Release}
\label{app:factnet_appendix}

This section supplements Section~\ref{sec:factnet_construction} with the deterministic implementation specifications required to reconstruct FactNet from Wikimedia dumps. The procedures detailed below are strictly non-stochastic and rely exclusively on versioned inputs, parsers, and configurations recorded in the build manifest (Appendix~\ref{app:reproducibility}).

\subsection{Reproducibility Manifest and Build Configuration}
\label{app:reproducibility}

FactNet is designed as a hermetic, snapshot-conditioned resource: the validity of every identifier, offset, and derived edge is strictly bound to a specific configuration of input data and processing logic.

\paragraph{Immutable Input Specification.}
To guarantee byte-level reproducibility, the manifest strictly pins all upstream dependencies. First, it records the exact Data Provenance by storing the URLs, timestamps, and SHA-256 checksums for the Wikidata JSON dump and all relevant Wikipedia XML and SQL dumps. Second, it locks the Execution Environment: since evidence pointers rely on consistent text segmentation and AST parsing, the manifest records the Git commit hash of the builder code and the container image digest. This ensures that system-level dependencies, such as ICU libraries\footnote{\url{https://icu.unicode.org}} for Unicode normalization, remain constant across reconstructions.

\paragraph{Versioned Policy and Configuration.}
The manifest explicitly versions the artifacts that control auditable canonicalization to prevent semantic drift. It includes cryptographic hashes for the normalization policy $\pi$, language packs, and the relation map. Any modification to these policies necessitates a new build identifier, ensuring that the semantic criteria used to merge statements or infer edges are traceable to a specific configuration version.

\subsection{FactSense Extraction Specification}
\label{app:factnet_factsense}

This subsection specifies the deterministic pipeline for aligning atomic statements $f=(S,P,V,Q)$ to re-locatable evidence units within a specific Wikipedia edition $\ell$. The process guarantees that every extraction decision is reproducible from the released snapshot and that all span offsets are stable relative to the canonicalized views described in Section~\ref{sec:offsets_main}.

\textbf{Subject Page Resolution and Scope.}
Extraction is strictly confined to the scope of a single subject page $p_\ell(S)$ per entity to ensure provenance clarity. For a target language $\ell$, the pipeline first attempts to resolve $S$ via explicit Wikidata sitelinks. If no sitelink exists, an optional conservative fallback mechanism generates a candidate title using a language-specific normalization function $\mathrm{TitleNorm}_{\ell}(\cdot)$. This fallback accepts a page if and only if the normalized title resolves to exactly one Namespace-0 page in the snapshot, rejecting any ambiguous redirects or collision-prone titles. Pages identified as disambiguation pages are systematically excluded.

\textbf{Canonical View Construction.}
To maintain offset stability, we eschew full template expansion because it introduces dependencies on transcluded resources. Instead, we generate three provenance-stable views, Sentence, Template, and Table, using a deterministic AST parser. All views undergo a uniform normalization $\mathrm{NormText}_{\ell}(\cdot)$, which applies Unicode NFC normalization, canonicalizes directional marks, and collapses whitespace while preserving semantic separators. Sentence views are derived by stripping markup and segmenting text via Stanza~\citep{qi-etal-2020-stanza} or rule-based splitters. Template views extract parameters from authorized infobox patterns without recursive rendering. Table views linearize cell content into coordinates comprising the table identifier, row, and column. All span offsets are defined as half-open intervals on the Unicode codepoints of these normalized strings, ensuring exact re-locatability.

\textbf{Hierarchical Matching Logic.}
Candidate evidence is identified through a prioritized hierarchy of matchers. Structure-based matching aligns literal values within infobox parameters or table cells using a versioned schema map; values are normalized via policy $\pi$ before comparison. Link-based matching handles entity-valued statements by resolving Wikilinks in the text to QIDs; resolution follows a strict precedence of direct page matches, then deterministic redirect chains, and finally unique normalized title matches. Links are accepted only if the resolved QID matches $V$ exactly. Lexical matching aligns literals in prose using strict, type-aware parsers. We explicitly avoid fuzzy matching to preserve auditability: dates must parse deterministically under the locale of the language pack, and quantities must satisfy unit constraints defined in $\pi$. Candidates are deduplicated by evidence unit, prioritizing Structure over Link, and Link over Lexical matches.

\textbf{Deterministic Confidence Scoring.}
Each retained FactSense is assigned a confidence score $c \in [0.5, 0.95]$, computed not as a probability but as a monotonic indicator of extraction precision. The score is calculated as:

\begin{equation}
    c = \mathrm{clip}\left( c_{\texttt{base}}(\tau) \cdot c_{\texttt{dtype}}(P,V) \cdot c_{\texttt{resolve}} \cdot c_{\texttt{amb}} \cdot c_{\texttt{sanity}} \right)
, \end{equation}

Here, $c_{\texttt{base}}(\tau)$ reflects the prior reliability of the match type $\tau$. The factors $c_{\texttt{resolve}}$ and $c_{\texttt{amb}}$ penalize indirect resolution and local ambiguity, respectively. The $c_{\texttt{sanity}}$ factor enforces datatype-specific invariants. All parameters are versioned in the build configuration, allowing users to reconstruct or recalibrate scores independently.

\subsection{Canonical Schema and Deterministic Identifiers}
\label{app:factnet_schema}

We enforce a strict separation between externally anchored identifiers, which are inherited from source dumps, and content-derived identifiers, which are computed via deterministic hashing of canonicalized data. This distinction ensures that the resource remains auditably reconstructible.

\paragraph{Canonical Serialization.}
To guarantee exact reproducibility across computing platforms, all records serve as inputs to a canonical JSON serialization protocol adhering to the RFC~8785 standard~\citep{rundgren2020rfc}. Object keys are sorted alphabetically by Unicode code point, insignificant whitespace is eliminated, and numeric values use the shortest-roundtrip representation. String fields derived from text processing, such as page titles or evidence snippets, are normalized to Unicode Normalization Form C\footnote{\url{https://unicode.org/reports/tr15}} (NFC). Conversely, inherited identifiers, including Wikidata QIDs and dump timestamps, are preserved verbatim to maintain referential integrity with the source infrastructure.

\paragraph{Deterministic Identifier Construction.}
Content-derived identifiers are generated using a domain-separated hashing scheme. Let $\mathrm{canon}(x)$ denote the canonical JSON serialization of an object $x$. We define the identifier $H_\tau(x)$ for a domain type $\tau$ as follows:
\begin{equation}
    H_\tau(x) = \texttt{SHA1}\left( \tau \parallel \texttt{0x1F} \parallel \texttt{build\_id} \parallel \texttt{0x1F} \parallel \mathrm{canon}(x) \right),
\end{equation}
Here, $\parallel$ denotes string concatenation, \texttt{0x1F} represents the ASCII Unit Separator, and \texttt{build\_id} corresponds to the snapshot version. This construction ensures that identifiers remain stable for identical contents within a snapshot while remaining distinct across incompatible builds. SHA-1 is selected for computational efficiency, as cryptographic collision resistance against adversarial inputs is not a design constraint for this static resource.

\paragraph{FactStatement Schema.}
FactStatements represent atomic assertions anchored by stable Wikidata identifiers. As detailed in Table~\ref{tab:schema_factstatement}, the schema preserves the original data topology, including qualifiers, references, and ranks. It augments the raw data with a \texttt{claim\_hash} for preliminary deduplication and pre-computed retrieval metadata to facilitate downstream grounding.

\begin{table}[h]
    \centering
    \caption{Schema definition for \textbf{FactStatement}. Key: \texttt{statement\_id} (External).}
    \label{tab:schema_factstatement}
    \small
    \begin{tabular}{p{0.25\textwidth} p{0.12\textwidth} p{0.55\textwidth}}
    \toprule
    \textbf{Field} & \textbf{Type} & \textbf{Description} \\
    \midrule
    \texttt{statement\_id} & String & Primary key (Wikidata Statement ID). \\
    \texttt{subject\_qid}, \texttt{property\_pid} & String & Entity and Property identifiers ($S, P$). \\
    \texttt{value} & Object & Typed value payload ($V$) preserving Wikidata datatype. \\
    \texttt{qualifiers} & Map & Qualifier multiset $Q$ mapped as \texttt{PID $\to$ [Value]}. \\
    \texttt{rank} & Enum & Rank status: \texttt{preferred}, \texttt{normal}, or \texttt{deprecated}. \\
    \texttt{references} & List & Raw reference objects preserving source provenance. \\
    \texttt{confidence} & Float & Heuristic score derived from rank and reference count. \\
    \texttt{sitelinks} & Map & Multilingual page title mapping: \texttt{lang $\to$ title}. \\
    \texttt{claim\_hash} & String & Hash of $(S, P, \mathrm{Norm}(V), \mathrm{Norm}(Q))$ for fast grouping. \\
    \bottomrule
    \end{tabular}
\end{table}

\paragraph{FactSense Schema.}
FactSenses represent grounded textual evidence. Unlike statements, FactSenses utilize content-derived keys generated from the tuple comprising the statement identifier, page identifier, and evidence pointer. This ensures that identical extractions yield stable identifiers regardless of pipeline execution order. The \texttt{evidence\_pointer} uniquely locates the span using structural indices, such as sentence index or template path, rather than brittle byte offsets. This design maximizes resilience to minor parser variations (Table~\ref{tab:schema_factsense}).

\begin{table}[h]
    \centering
    \caption{Schema definition for \textbf{FactSense}. Key: Content-derived $H_{\texttt{factsense}}(\cdot)$.}
    \label{tab:schema_factsense}
    \small
    \begin{tabular}{p{0.25\textwidth} p{0.12\textwidth} p{0.55\textwidth}}
    \toprule
    \textbf{Field} & \textbf{Type} & \textbf{Description} \\
    \midrule
    \texttt{factsense\_id} & String & Unique hash of the alignment instance. \\
    \texttt{statement\_id} & String & Foreign key to the supported FactStatement. \\
    \texttt{language}, \texttt{page\_id} & String/Int & Wikipedia edition code and Page ID. \\
    \texttt{evidence\_pointer} & Object & Deterministic locator (e.g., \texttt{\{unit\_type: SENTENCE, index: 4\}}). \\
    \texttt{sentence} & String & The raw text span containing the evidence. \\
    \texttt{match\_type} & Enum & Alignment strategy (e.g., \texttt{sitelink}, \texttt{infobox\_kv}). \\
    \texttt{confidence} & Float & Alignment confidence score $[0.5, 1.0]$. \\
    \texttt{provenance} & Object & Extraction metadata (timestamp, parser version). \\
    \bottomrule
    \end{tabular}
\end{table}

\paragraph{FactSynset Schema.}
FactSynsets aggregate semantically equivalent statements. The identifier is derived from an \texttt{aggregation\_key} constructed by normalizing values and qualifiers under policy $\pi$. To support auditability, any relaxation of strict equivalence, such as unit conversion or precision truncation, requires explicit documentation in the \texttt{merge\_reasons} field (Table~\ref{tab:schema_synset}).

\begin{table}[h]
    \centering
    \caption{Schema definition for \textbf{FactSynset}. Key: Content-derived $H_{\texttt{synset}}(\texttt{agg\_key})$.}
    \label{tab:schema_synset}
    \small
    \begin{tabular}{p{0.25\textwidth} p{0.12\textwidth} p{0.55\textwidth}}
    \toprule
    \textbf{Field} & \textbf{Type} & \textbf{Description} \\
    \midrule
    \texttt{synset\_id} & String & Unique hash of the aggregation key. \\
    \texttt{aggregation\_key} & String & Canonical form of $S \parallel P \parallel \mathrm{Norm}(V) \parallel \mathrm{Norm}(Q)$. \\
    \texttt{member\_statement\_ids} & List & List of aggregated FactStatement IDs. \\
    \texttt{canonical\_mentions} & Map & Best multilingual evidence: \texttt{lang $\to$ \{factsense\_id, ...\}}. \\
    \texttt{merge\_reasons} & List & Justifications for aggregation (e.g., \texttt{value\_normalization}). \\
    \texttt{aggregate\_confidence} & Float & Aggregated confidence score (max of members). \\
    \bottomrule
    \end{tabular}
\end{table}

\paragraph{RelationEdge Schema.}
Edges represent computed relationships between FactSynsets. To differentiate between edges induced by distinct logical rules, the identifier incorporates the specific rule identifier. This allows users to trace any edge back to the specific mapping file or heuristic that generated it (Table~\ref{tab:schema_relation}).

\begin{table}[h]
    \centering
    \caption{Schema definition for \textbf{RelationEdge}. Key: Content-derived $H_{\texttt{relation}}(\cdot)$.}
    \label{tab:schema_relation}
    \small
    \begin{tabular}{p{0.25\textwidth} p{0.12\textwidth} p{0.55\textwidth}}
    \toprule
    \textbf{Field} & \textbf{Type} & \textbf{Description} \\
    \midrule
    \texttt{relation\_id} & String & Unique hash of endpoints and rule. \\
    \texttt{source\_synset\_id} & String & Source FactSynset ID. \\
    \texttt{target\_synset\_id} & String & Target FactSynset ID. \\
    \texttt{relation\_type} & String & Relation category (e.g., \texttt{temporal\_before}, \texttt{equivalent}). \\
    \texttt{rule\_id} & String & Identifier of the rule or mapping generating the edge. \\
    \texttt{evidence} & Object & Supporting metadata (e.g., source counts, intermediate keys). \\
    \bottomrule
    \end{tabular}
\end{table}

\subsection{Normalization Policy $\pi$ and Claim Hashing}
\label{app:factnet_normalization}

This section details the versioned normalization policy $\pi$ governing FactSynset construction. We define a Wikidata statement as a tuple $f = (S, P, V, Q)$, where $S$ is the subject QID, $P$ is the property PID, $V$ is the main snak value, and $Q$ is a multiset of qualifier snaks. The policy $\pi$ ensures that statement canonicalization is deterministic and auditable. It is distributed as a machine-readable configuration containing datatype defaults, property-specific overrides, and allowlists for semantic relaxations.

\paragraph{Value Normalization ($\mathrm{NormValue}$).}
The function $\mathrm{NormValue}(P, V; \pi)$ maps typed values to a canonical serialized form. We handle Wikidata snak types \texttt{novalue} and \texttt{somevalue} by mapping them to reserved constants to avoid collision with literals. For standard values, the normalization logic is datatype-specific. Entities are mapped to canonical QID strings; we strictly avoid redirect resolution at this layer to prevent semantic drift, as aliasing is handled exclusively during FactSense grounding. Quantities and coordinates are parsed into high-precision decimal representations to ensure platform independence. Unit conversion and coordinate rounding are disabled by default and occur only if explicitly authorized by $\pi$ for specific properties. Temporal values are serialized to ISO-8601\footnote{\url{https://en.wikipedia.org/wiki/ISO_8601}} strings with explicit precision attributes; relaxation, such as truncating days to months, is applied only via allowlist gating. Finally, string values undergo Unicode NFC normalization and whitespace trimming, while aggressive stemming or case-folding is disabled unless specified per-property.

\paragraph{Order-Invariant Qualifier Normalization ($\mathrm{NormQuals}$).}
As Wikidata qualifiers are unordered multisets, we define $\mathrm{NormQuals}(Q; \pi)$ to guarantee deterministic serialization. For each qualifier snak $(p_q, v_q) \in Q$, we compute the normalized pair $(p_q, \mathrm{NormValue}(p_q, v_q; \pi))$. The resulting list of pairs is sorted lexicographically by the tuple key. This renders the aggregation key invariant to the input order of qualifiers.

\paragraph{Claim Hashing and Aggregation.}
To establish FactSynset membership, we construct a strict \texttt{aggregation\_key} by concatenating the canonical serializations of the statement components: $S \parallel P \parallel \mathrm{NormValue}(V) \parallel \mathrm{NormQuals}(Q)$. For indexing efficiency, we compute a compact \texttt{claim\_hash} using SHA-256 applied to the UTF-8 bytes of the key. FactNet treats the hash solely as an index bucket, meaning full key equality is verified before merging records.

\paragraph{Auditable Merge Provenance.}
FactNet distinguishes between strict merges, resulting from lossless normalization such as whitespace trimming, and relaxed merges resulting from information-reducing transformations. Any application of a relaxed policy, such as \texttt{TIME\_PRECISION\_RELAX} or \texttt{UNIT\_CONVERT}, generates a structured \texttt{merge\_reason} token stored in the Synset metadata. This mechanism allows downstream users to filter the knowledge graph based on the strictness of semantic equivalence.

\subsection{FactSynset Construction and Canonical Selection}
\label{app:factnet_synsets}

This section details the deterministic procedure mapping atomic FactStatements into FactSynset equivalence classes and the selection of canonical representatives. The process relies exclusively on a versioned normalization policy $\pi$ and fixed Wikimedia dumps to ensure full auditability.

\paragraph{Aggregation and Normalization Policy.}
A FactSynset is defined as the equivalence class of FactStatements sharing an identical aggregation key. For a statement $f=(S,P,V,Q)$, where $Q$ is a multiset of qualifier pairs, the key is constructed as:
\begin{equation}
    \mathcal{K}(f; \pi) = S \parallel P \parallel \mathrm{NormVal}(P, V; \pi) \parallel \mathrm{NormQuals}(Q; \pi).
\end{equation}
The function $\mathrm{NormVal}$ applies datatype-specific normalization strictly regulated by the policy $\pi$ via per-property allowlists and thresholds. To guarantee identifier stability against JSON serialization variances, $\mathrm{NormQuals}$ normalizes individual qualifiers and sorts them deterministically by the tuple comprising the PID and normalized value prior to hashing. The resulting \texttt{synset\_id} is a cryptographic hash of $\mathcal{K}$ concatenated with the policy version ID, ensuring that any modification to equivalence criteria yields distinct identifiers.

\paragraph{Canonical Statement Selection.}
To facilitate inspection, each synset designates a single \texttt{canonical\_statement\_id}, $\hat{f}$, selected to maximize authority signals. Let $\rho(f) \in \{\texttt{preferred}, \texttt{normal}, \texttt{deprecated}\}$ be the Wikidata rank, $\mathcal{R}(f)$ be the count of distinct reference blocks, and $t(f)$ be the last edit timestamp. We define a scoring tuple and select $\hat{f}$ via lexicographical maximization:
\begin{equation}
    \hat{f} = \operatorname*{arg\,max}_{f \in \text{Synset}} \Big( \mathbb{I}_{\rho}[\rho(f)],\ \mathcal{R}(f),\ t(f),\ -\texttt{id}(f) \Big),
\end{equation}
Here, $\mathbb{I}_{\rho}$ maps ranks to monotonic integer scores and $-\texttt{id}(f)$ serves as a deterministic tie-breaker. This selection requires no learned parameters and strictly favors statements that are editor-preferred and well-referenced.

\paragraph{Canonical Mention Selection (FactSense).}
For each language $\ell$, we identify a canonical mention from the pool of FactSenses associated with the synset's members. We first deduplicate mentions by their evidence pointer to avoid over-counting redundant matches. Candidates are then ranked by a hierarchy of evidence reliability, preferring infobox fields over table cells, and table cells over sentences. Within unit types, we prioritize link-based matching over lexical matching. Final ties are broken by confidence score and pointer stability. This ensures that the canonical mention points to the most structured and unambiguous evidence available for the fact in language $\ell$.

\subsection{Re-locatable Evidence Pointers and Offset Computation}
\label{app:factnet_offsets}

This subsection details the mechanism used to ground FactStatements to concrete, reproducible spans within Wikipedia. To ensure auditability and immunity to the dynamic nature of online content, FactNet defines evidence pointers relative to specific dump snapshots and deterministic processing pipelines rather than unstable live URLs.

\paragraph{Pointer Schema and Scope.}
A FactSense pointer corresponds to the tuple:
\begin{equation}
    \mathcal{P} = (\texttt{page\_id}, \texttt{revision\_id}, \texttt{view}, \texttt{locator}, \texttt{start}, \texttt{end}, \texttt{norm\_id}).
\end{equation}
The \texttt{page\_id} and \texttt{revision\_id} refer to the specific MediaWiki XML dump snapshot recorded in the build manifest. The \texttt{view} and \texttt{locator} jointly isolate a discrete evidence unit, while \texttt{start} and \texttt{end} define a character span within that unit. Crucially, these coordinates are valid only within the context of the versioned preprocessing configuration identified by \texttt{norm\_id}, ensuring that changes in segmentation logic or text normalization do not silently invalidate offsets.

\paragraph{Deterministic Views and Unit Locators.}
To locate evidence without relying on byte offsets in the raw XML, we define three provenance-stable views. The \texttt{locator} syntax depends on the selected view. For the Sentence View, the locator is an integer index representing the sentence's position in the sequence generated by our deterministic pipeline. For the InfoBox View, the locator is a composite key consisting of the template path and parameter name, where the path disambiguates repeated templates via traversal order. For the Table View, the locator is a tuple $(i, r, c)$ specifying the $i$-th wikitable and the grid coordinates after resolving row and column spans.

\paragraph{Normalization and Offset Definition.}
Offsets are computed on a normalized evidence string $\tilde{x}$ rather than raw wikitext. Let $x$ be the raw string extracted from the locator. We apply a normalization function $\tilde{x} = \mathcal{N}_{\texttt{norm\_id}}(x)$ which enforces Unicode NFC normalization, standardizes newlines, removes zero-width characters, and deterministically decodes a bounded set of HTML entities. Whitespace handling is view-specific: maximal collapsing is applied to sentences, while structure-preserving normalization is used for tables. The span indices $[\texttt{start}, \texttt{end})$ represent Unicode codepoint offsets in $\tilde{x}$. This abstraction shields the dataset from implementation-specific byte encoding differences across programming languages.

\paragraph{Reconstruction Protocol.}
Re-locating a span follows a deterministic procedure: retrieve the raw page content using the page and revision identifiers, regenerate the specific view structure using the released parser versions, select the unit via the locator, apply the normalization function $\mathcal{N}_{\texttt{norm\_id}}$, and slice the string using the codepoint offsets. This protocol allows users to verify the exact textual evidence used during construction without distributing the full text of Wikipedia, ensuring compliance with licensing attribution requirements while maximizing reproducibility.

\subsection{Multilingual Language Packs and Sentence Segmentation}
\label{app:factnet_languagepacks}

To ensure strict reproducibility across 316 languages, FactNet eliminates hidden degrees of freedom in text processing via Language Packs. These are versioned, machine-readable specifications that fully determine the mapping from raw page text to sentence units. Unlike pipelines that rely on implicit library defaults or locale-dependent heuristics, our language packs explicitly define the segmentation backend, text normalization rules, and boundary exceptions for each Wikipedia edition $\ell$.

\paragraph{Specification and Versioning.}
A language pack is serialized as a canonical JSON object containing all parameters necessary to reproduce segmentation deterministically. To guarantee traceability, every FactSense record references a \texttt{language\_pack\_id}, computed as the SHA-256 hash of the pack's content. This mechanism creates a cryptographic binding between the dataset and the processing logic, ensuring that segmentation decisions remain reconstructible even if upstream libraries update their default behaviors. Table~\ref{tab:language_pack_fields} summarizes the core configuration fields.

\begin{table}[h]
\centering
\small
\caption{Core components of a FactNet Language Pack. All fields are versioned and included in the \texttt{language\_pack\_id} hash to guarantee deterministic reconstruction.}
\label{tab:language_pack_fields}
\begin{tabular}{lp{0.75\linewidth}}
\toprule
\textbf{Component} & \textbf{Deterministic Functionality} \\
\midrule
\texttt{backend} & Engine selection (\texttt{stanza} or \texttt{rule\_based}) with pinned version strings. \\
\texttt{model\_id} & Fully qualified model identifier and checksum (for Stanza backends). \\
\texttt{normalization} & Unicode form (e.g., NFKC) and whitespace policies applied pre-segmentation. \\
\texttt{terminal\_punct} & Set of sentence-final characters $\mathcal{P}_{\ell}$ (for rule-based backends). \\
\texttt{suppression} & Paired delimiters $\mathcal{D}_{\ell}$ (brackets, quotes) and abbreviation exceptions. \\
\texttt{wiki\_rules} & Rules for title normalization and disambiguation logic specific to $\ell$. \\
\bottomrule
\end{tabular}
\end{table}

\paragraph{Deterministic Normalization and Offsets.}
Consistency in offsets requires a stable coordinate system. Let $x$ be the provenance-stable plain text derived from the wikitext dump. The language pack defines a normalization function $N_{\ell}$, handling Unicode normalization and whitespace canonicalization, to produce a normalized evidence string $\tilde{x} = N_{\ell}(x)$. All segmentation operations operate on $\tilde{x}$, and the resulting sentence boundaries $[b_k, e_k)$ are stored as Unicode codepoint indices relative to $\tilde{x}$. This ensures that evidence pointers remain valid across different hardware and platforms.

\paragraph{Segmentation Backends.}
FactNet supports two execution paths, both strictly governed by the pack configuration. For high-resource languages, we employ Stanza. To mitigate non-determinism, the language pack pins the exact library version and model artifact checksum. At runtime, the pipeline disables GPU acceleration and multi-threading, and strictly enforces the tokenizer configuration specified in the pack. For low-resource languages or where explicitly configured, we use a deterministic scanner. Candidate boundaries defined by terminal punctuation $\mathcal{P}_{\ell}$ are filtered through a suppression stack that tracks paired delimiters to prevent splitting nested clauses. Additionally, context-aware exception patterns and minimum-length constraints are applied to prioritize precision.

\subsection{Deterministic Derivation of RelationEdges}
\label{app:factnet_relations}

We define the set of RelationEdges, $\mathcal{E}$, as the output of a pure, deterministic function acting on the set of FactSynsets, a set of versioned rule artifacts, and the global configuration $\pi$. Unlike edges in probabilistic knowledge graphs, FactNet relations are not learned but strictly derived from rule satisfaction, ensuring that any user with the released build manifest can reproduce the edge set.

\textbf{Derivation Operators.}
We employ three operators to instantiate edges. First, Direct Joins link entity-valued synsets to the descriptive context of that entity. For a synset $y$ where the normalized value resolves to an entity object $O_y$, we generate edges $y \to z$ for all target synsets $z$ where the subject of $z$ is $O_y$, subject to property allowlists. Second, Schema Mappings utilize the \texttt{PROPERTY\_RELATION\_MAP}. For each row mapping a PID $P$ to a relation type $R$ with constraints $C$, if $y$ satisfies these constraints, we emit an edge to targets defined by the mapping logic. Third, Bounded Traversal approximates relations requiring intermediate hops. If authorized by the map, we search for valid paths in the graph. The full path of intermediate synset IDs is recorded in the edge provenance to maintain auditability.

\textbf{Conflict and Signal Generation.}
FactNet treats logical inconsistencies as informational signals rather than grounds for deletion. We derive \texttt{POTENTIAL\_CONFLICT} edges through two mechanisms: functional violations and temporal overlap. Functional violations occur when properties restricted as functional link synsets sharing the same subject and property but possessing distinct normalized values. Temporal overlap violations occur if synsets violate a functional constraint while their temporal intervals overlap. These edges allow downstream systems to filter or analyze contested facts without altering the underlying atomic assertions.

\textbf{Aggregation and Confidence Scoring.}
Duplicate derivations are aggregated by maximizing confidence and concatenating evidence traces. The confidence score of an edge $e$ is computed as a scalar in $[0,1]$:
\begin{equation}
    \texttt{conf}(e) = \min\left(1, \; w_{tier} \cdot c(y) \cdot \Psi_{ref}(y) \cdot \Psi_{lang}(y) \right),
\end{equation}
Here, $w_{tier}$ is the released weight of the derivation rule, $c(y)$ is the source synset's aggregate confidence, and $\Psi_{ref}, \Psi_{lang}$ are monotonic saturating functions of the source reference count and language coverage, respectively. This scoring policy prioritizes relations supported by highly corroborated, multilingual, and high-authority source facts.

\subsection{Release Organization, Formats, and Indexing}
\label{app:release_format}

To facilitate scalable analytics while ensuring strict auditability, the FactNet release is organized into deterministic, immutable artifacts. The distribution design prioritizes reproducibility: given a fixed build context, the sharded outputs are byte-for-byte reconstructible.

\paragraph{Formats and Schema Versioning.}
We release synchronized representations of all record families in two formats: JSONL for auditing and Parquet\footnote{\url{https://parquet.apache.org}} for high-throughput analytics. Both formats adhere to strict schemas versioned alongside the dataset. A top-level manifest records the cryptographic checksums of all shards, the schema version, and the upstream Wikimedia dump identifiers utilized in the build. Nested structures are typed explicitly to prevent ambiguous string flattening.

\paragraph{Deterministic Sharding Protocol.}
Data partitions are generated via a stateless hashing protocol ensuring uniform distribution and parallel processing capability. For a record $r$ with a stable primary identifier $\mathrm{id}(r)$, the shard assignment is defined as $\mathrm{shard}(r) = \mathcal{H}(\mathrm{id}(r)) \bmod N$, where $N$ is the fixed partition count and $\mathcal{H}$ is a seeded 64-bit hash function. Within each shard, records are sorted by identifier to maximize compression ratios and ensure deterministic file output.

\subsection{Licensing and Evidence-Text Packaging}
\label{app:licensing}

FactNet utilizes a decoupled distribution architecture to strictly adhere to Wikimedia licensing constraints while maximizing utility for graph-based research.
Wikidata content is released under CC0\footnote{\url{https://creativecommons.org/publicdomain/zero/1.0/}}, whereas Wikipedia textual content inherits the CC BY-SA license\footnote{\url{https://creativecommons.org/licenses/by-sa/4.0/}}.
To operationalize this distinction, we partition the dataset into two artifacts: a default Structural Pack and an optional Evidence-Text Pack.

\paragraph{Default Distribution: Structural Pack (CC0).}
The core FactNet release contains the complete graph topology and grounding metadata but excludes all expressive text strings. Instead, it relies on the reconstructible pointer mechanism defined in Section~\ref{sec:offsets_main}. Each FactSense record stores a provenance tuple and deterministic character offsets relative to a normalized view. This design ensures that the artifact remains lightweight and permissible for topology-centric research without triggering ShareAlike obligations.

\paragraph{Optional Distribution: Evidence-Text Pack (CC BY-SA).}
For applications requiring immediate access to textual evidence, we provide an opt-in bundle containing the pre-computed, normalized evidence strings. This pack is distributed under CC BY-SA and includes mandatory attribution metadata—such as source snapshots, page titles, and revision identifiers—embedded within each record to facilitate compliance. We also include deterministic checksums for each normalized string to verify consistency with the versioned language packs used during construction.
\section{Extended Statistics and Quality Assessment Details}
\label{app:stats_appendix}

This section provides comprehensive assessments, detailed distributional breakdowns, and the full audit protocol supporting the top-level statistics reported in \S\ref{sec:stats_quality}.
Consistent with the main text, all analyses correspond to the \texttt{2025-11-01} snapshot build. Unless noted otherwise, counts are computed after deduplication at the released primary keys (specifically \texttt{sense\_id}, \texttt{statement\_id}, and \texttt{synset\_id}) using the default build configuration described in \S\ref{sec:stats_quality}.
When reporting percentages over FactSenses, we weight by the number of FactSenses rather than by pages or entities; conversely, when reporting percentages over synsets, we weight by FactSynsets rather than by individual statements.

\subsection{Distributional Diagnostics and Coverage Strata}
\label{app:summary_full}

Although FactNet operates at substantial scale, the distribution of evidence is naturally skewed by the underlying data availability in Wikidata and Wikipedia. We analyze these distributions to ensure transparency regarding the long-tail behavior that affects benchmarking, sampling, and downstream training stability.

\noindent\textbf{Language Tiers.}
To facilitate stratified analysis, we categorize the 316 supported languages into three tiers based on their Wikipedia article count ($N_{\text{art}}$) at the time of the snapshot. \emph{High-Resource (Tier 1)} languages are defined by $N_{\text{art}} \ge 100\text{K}$, comprising 71 languages that cover 84.3\% of FactSenses. \emph{Medium-Resource (Tier 2)} languages are defined by $10\text{K} \le N_{\text{art}} < 100\text{K}$, comprising 94 languages that cover 12.8\% of FactSenses. \emph{Low-Resource (Tier 3)} languages are defined by $N_{\text{art}} < 10\text{K}$, comprising 151 languages that cover 2.9\% of FactSenses. Although Tier 3 contributes a small fraction of the total volume, it represents nearly half of the linguistic diversity. Grounding precision remains stable across tiers, as detailed in Appendix~\ref{app:grounding_audit}, though recall declines in Tier 3 owing to shorter page lengths, fewer standardized infobox templates, and reduced lexical redundancy.

\noindent\textbf{Language long tail and concentration.}
To operationalize the long tail for benchmarking, we report additional concentration measures. Let $x_\ell$ denote the number of FactSenses in language $\ell$, and let $p_\ell = x_\ell / \sum_{\ell'} x_{\ell'}$. We compute a Gini coefficient over $\{p_\ell\}$ ($G \approx 0.91$) and the effective number of languages $N_{\text{eff}} = \exp(H)$, where $H = -\sum_\ell p_\ell \log p_\ell$ is the Shannon entropy ($N_{\text{eff}} \approx 18.7$). These values indicate that although 316 languages are covered, the distributional support is comparable to only a few dozen equally represented languages. This finding motivates our use of stratified evaluation and tier-wise reporting.

\begin{table}[h]
\centering
\caption{\textbf{Top-language FactSense concentration.} Shares are computed over all FactSenses. The table is intended to accompany the language-rank CDF in Fig.~\ref{fig:language_cdf}.}
\label{tab:top_languages}
\small
\setlength{\tabcolsep}{6pt}
\begin{tabular}{l r r}
\toprule
\textbf{Language} & \textbf{FactSenses (B)} & \textbf{Share (\%)} \\
\midrule
English (\texttt{en}) & 0.53 & 17.6 \\
German (\texttt{de}) & 0.24 & 8.0 \\
French (\texttt{fr}) & 0.21 & 7.0 \\
Spanish (\texttt{es}) & 0.18 & 6.0 \\
Russian (\texttt{ru}) & 0.16 & 5.3 \\
Italian (\texttt{it}) & 0.12 & 4.0 \\
Japanese (\texttt{ja}) & 0.11 & 3.7 \\
Portuguese (\texttt{pt}) & 0.10 & 3.3 \\
Chinese (\texttt{zh}) & 0.09 & 3.0 \\
Polish (\texttt{pl}) & 0.08 & 2.7 \\
\midrule

\textbf{Top-5 total} & \textbf{1.32} & \textbf{43.9} \\
\textbf{Top-10 total} & \textbf{1.82} & \textbf{60.6} \\
\bottomrule
\end{tabular}
\end{table}

\begin{figure}[t]
\centering
\includegraphics[width=0.75\textwidth]{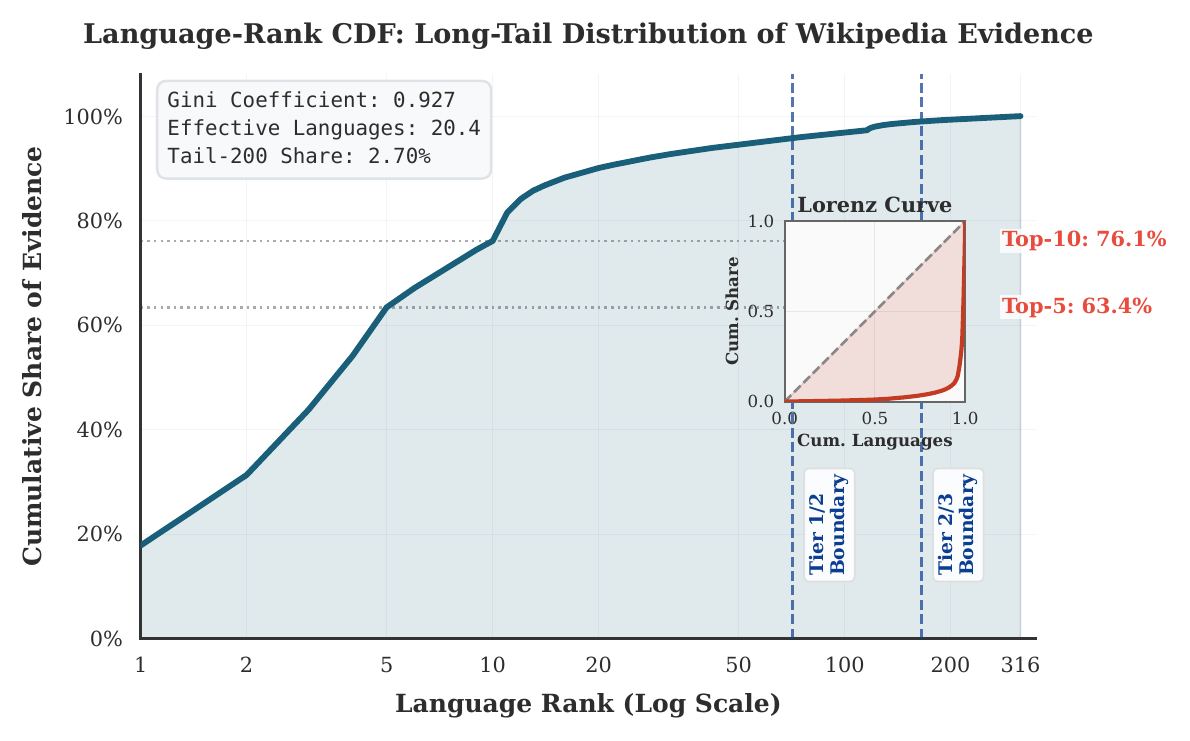}
\caption{\textbf{Language-rank distribution diagnostics.} This figure complements Table~\ref{tab:top_languages} and justifies stratified sampling and tier-wise reporting.}
\label{fig:language_cdf}
\end{figure}

\noindent\textbf{Evidence unit composition.}
Because FactSenses can be grounded in sentences, infobox fields, or table cells, we provide a breakdown of evidence-unit types and their interaction with match mechanisms. This distinction is critical for benchmark construction, as models trained on sentence-only supervision may exhibit different behaviors compared to those trained on semi-structured evidence. Table~\ref{tab:evidence_unit_mix} reports a global aggregate. We also release the same table disaggregated by language tier and subject top-level type in Appendix~\ref{app:representation_bias}.

\begin{table}[t]
\centering
\caption{\textbf{Evidence-unit mix and match-mechanism composition.} Shares sum to 100\% within each panel. ``Sentence'' includes lead and non-lead sentences; ``Table'' aggregates wikitable and list-table cells; ``Infobox'' corresponds to templated key-value fields after normalization.}
\label{tab:evidence_unit_mix}
\small
\setlength{\tabcolsep}{6pt}
\begin{tabular}{l r r}
\toprule
\textbf{Evidence unit type} & \textbf{FactSenses share (\%)} & \textbf{Strong-evidence share (\%)} \\
\midrule
Sentence & 57.5 & 49.2 \\
Infobox field & 28.4 & 45.8 \\
Table cell & 14.1 & 5.0 \\
\bottomrule
\end{tabular}

\setlength{\tabcolsep}{6pt}
\begin{tabular}{l r r}
\toprule
\textbf{Match type} & \textbf{FactSenses share (\%)} & \textbf{Precision target} \\
\midrule
\texttt{WIKILINK\_ENTITY} & 35.0 & High \\
\texttt{INFOBOX\_FIELD} & 20.0 & High \\
\texttt{LEXICAL\_VALUE} & 35.0 & Medium \\
\texttt{LEAD\_WEAK} & 10.0 & Lower \\
\bottomrule
\end{tabular}
\end{table}

\noindent\textbf{Synset multiplicity and canonicalization pressure.}
A FactSynset canonically groups one or more Wikidata statements judged equivalent under a strict or policy-relaxed equivalence relation, as defined in \S\ref{sec:factnet_construction}. For downstream applications, it is relevant whether a synset is typically a singleton (indicating low ambiguity) or the result of merging multiple statements (indicating higher canonicalization pressure). Table~\ref{tab:synset_multiplicity} summarizes the synset size distribution by the number of member statements. The ``policy-relaxed'' row counts synsets containing at least one member whose inclusion required a relaxation reason; these constitute a small subset and are explicitly traceable via \texttt{merge\_reasons}.

\begin{table}[t]
\centering
\caption{\textbf{FactSynset multiplicity diagnostics.} ``Size'' counts the number of member \texttt{statement\_id}s per synset.}
\label{tab:synset_multiplicity}
\small
\setlength{\tabcolsep}{6pt}
\begin{tabular}{l r r}
\toprule
\textbf{Synset size} & \textbf{Synsets (B)} & \textbf{Share (\%)} \\
\midrule
1 & 1.40 & 90.4 \\
2 & 0.11 & 7.1 \\
3--5 & 0.031 & 2.0 \\
$\ge 6$ & 0.009 & 0.6 \\
\midrule
Contains any policy-relaxed merge & 0.020 & 1.3 \\
\bottomrule
\end{tabular}
\end{table}

\noindent\textbf{Property Coverage CDF.}
FactNet covers 12.1K distinct properties, but evidence density varies substantially by property type. Table~\ref{tab:property_cdf} presents the Cumulative Distribution Function (CDF) of evidence-bearing synsets per property. Universal properties, such as \texttt{instance\_of}, \texttt{date\_of\_birth}, and \texttt{coordinate\_location}, saturate the head, whereas domain-specific identifiers and technical parameters populate the tail. For model developers, this implies that average performance on randomly sampled properties may be dominated by a small head unless property-balanced evaluation is explicitly employed.

\begin{table}[h]
\centering
\caption{\textbf{Property Coverage Distribution.} Cumulative count of properties exceeding specific evidence-bearing synset thresholds.}
\label{tab:property_cdf}
\small
\begin{tabular}{l r r p{0.45\linewidth}}
\toprule
\textbf{Min. Synsets} & \textbf{Count} & \textbf{\%} & \textbf{Example Properties} \\
\midrule
$\ge$ 10,000,000 & 18 & 0.15 & P31 (instance of), P21 (sex/gender), P131 (admin loc) \\
$\ge$ 1,000,000 & 142 & 1.17 & P57 (director), P577 (pub date), P856 (official website) \\
$\ge$ 100,000 & 583 & 4.81 & P2048 (height), P166 (award), P106 (occupation) \\
$\ge$ 10,000 & 1,139 & 9.40 & P212 (ISBN), P1619 (date of opening), P206 (inflows) \\
$\ge$ 1,000 & 3,852 & 31.80 & P1532 (country for sport), P1435 (heritage status) \\
$\ge$ 1 & 12,114 & 100.0 & \emph{Long-tail external IDs and technical specs} \\
\bottomrule
\end{tabular}
\end{table}

\noindent\textbf{Qualifier and reference density (statement-level diagnostics).}
Given that a core motivation for FactNet is to support provenance-aware benchmarking, we provide additional statement-level diagnostics beyond the aggregate proportions in Table~\ref{tab:at_a_glance}. For each property $P$, we compute the fraction of statements with at least one reference and the fraction with at least one qualifier. We then summarize these distributions across properties using percentiles to avoid overemphasizing head properties. The median property has 41\% of statements with references (10th percentile 12\%, 90th percentile 78\%) and 17\% with qualifiers (10th percentile 2\%, 90th percentile 49\%).

\subsection{The Evidence Gap: Funnel Analysis}
\label{app:funnel}

We quantify the evidence gap, defined as the discrepancy between facts present in Wikidata and those successfully grounded in Wikipedia, via a deterministic attribution funnel. We analyze the subset of FactSynsets where the subject has at least one Wikidata sitelink to a target language $\ell$. The funnel tracks retention through three primary stages. First, \emph{Page Retrieval} requires that the sitelink resolves to a valid, non-redirect, non-disambiguation page in the dump. Second, \emph{Unit Construction} requires that the parser extracts at least one valid evidence unit (sentence, infobox field, or table cell) after filtering empty pages and parse failures. Third, \emph{Matching} requires that the alignment pipeline identifies at least one FactSense for the fact within the scoped page content.
Table~\ref{tab:funnel_breakdown} reports macro-averaged retention rates across languages within each tier, where each language receives equal weight. We also compute a micro-average weighted by the number of sitelink-conditioned candidate synsets in each language. Micro-averages are consistently higher because high-resource languages exhibit higher yield and dominate volume (micro-average Matching Success: Tier 1 $=$ 0.82, Tier 2 $=$ 0.60, Tier 3 $=$ 0.39).

\begin{table}[h]
\centering
\caption{\textbf{Grounding Attribution Funnel.} Retention rates conditioned on the existence of a Wikidata sitelink. Values are macro-averaged across languages within each tier.}
\label{tab:funnel_breakdown}
\small
\begin{tabular}{l c c c}
\toprule
\textbf{Stage} & \textbf{Tier 1 (High)} & \textbf{Tier 2 (Med)} & \textbf{Tier 3 (Low)} \\
\midrule
1. Sitelink Exists (Condition) & 1.00 & 1.00 & 1.00 \\
2. Page Retrieval Success & 0.98 & 0.94 & 0.89 \\
3. Unit Construction Success & 0.96 & 0.91 & 0.82 \\
4. Matching Success ($\ge 1$ sense) & \textbf{0.79} & \textbf{0.58} & \textbf{0.36} \\
\midrule
\emph{Primary Loss Factor} & Matching & Matching & Page/Unit \\
\bottomrule
\end{tabular}
\end{table}

\noindent\textbf{Attribution of losses within stages.}
To make the funnel actionable for dataset users, we further decompose Page Retrieval failures into redirect-only sitelinks, disambiguation pages, and XML parsing errors. Similarly, we decompose Unit Construction failures into pages with only non-textual content under our renderer (such as galleries), pages with unsupported template constructs, and pages whose content falls entirely in excluded namespaces or sections. In Tier 3, a substantial portion of the Unit Construction drop is attributable to extremely short articles and list-like stubs that yield few extractable sentences after boilerplate removal. Specifically, 57\% of Unit Construction failures in Tier 3 are classified as ``stub/boilerplate dominated.''

\noindent\textbf{Match-stage bottlenecks and mitigation.}
Within the Matching stage, we identify two dominant bottlenecks. The first is alias coverage for entity values, particularly for scripts with rich orthographic variation or where Wikipedia favors localized exonyms while Wikidata aliases are sparse (48\% of Tier 3 match failures involve missing or low-recall alias generation). The second is template mapping coverage for infobox fields, where language-specific template keys are not fully mapped to Wikidata properties (36\% of Tier 2 match failures are attributable to missing template-key mappings).

\begin{figure}[t]
\centering
\includegraphics[width=\textwidth]{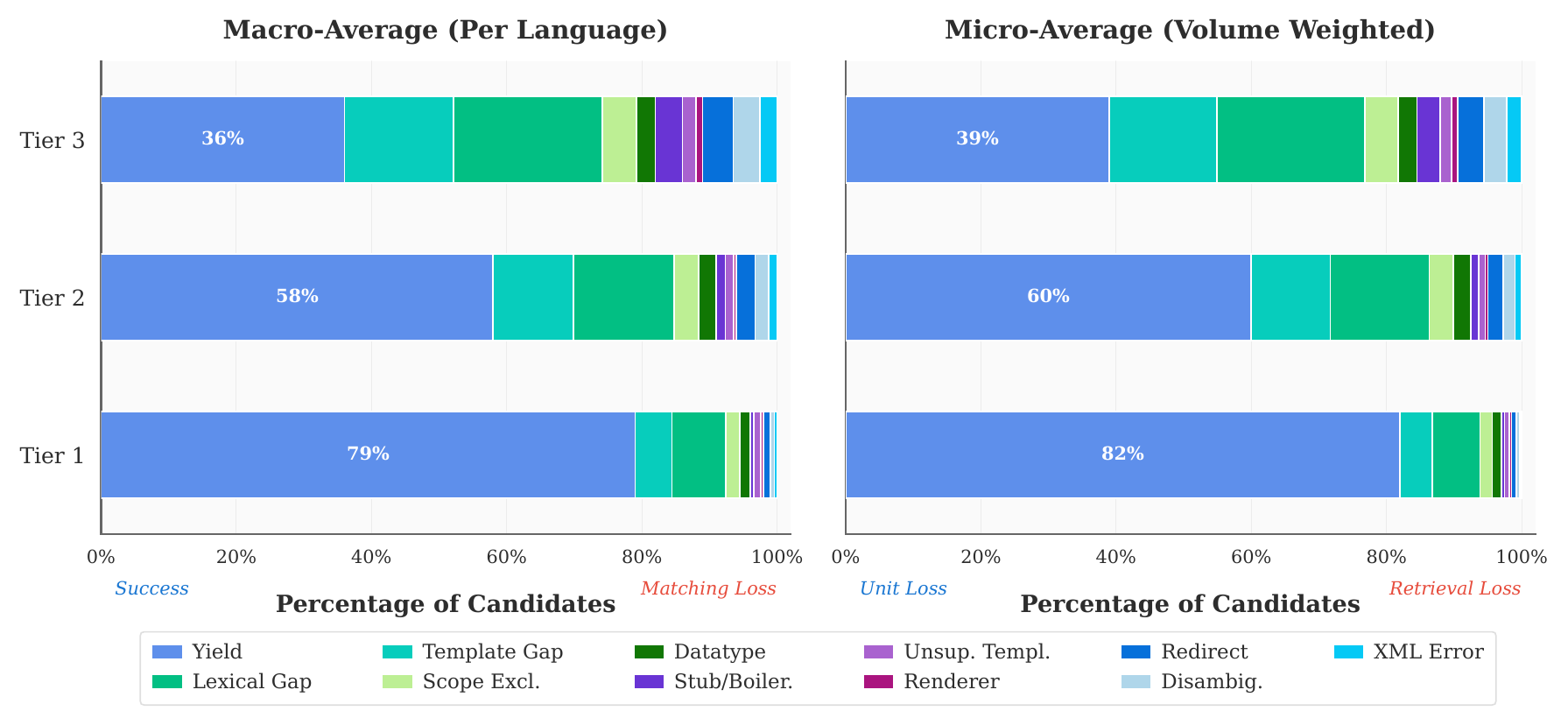}
\caption{\textbf{Evidence-gap funnel visualization.} We recommend plotting both macro- and micro-averaged funnels, and stacking the dominant loss reasons per tier. This figure operationalizes the ``evidence gap'' and helps users choose between improving extraction coverage versus restricting to the strong-evidence subset.}
\label{fig:funnel}
\end{figure}

\subsection{Representational Bias Diagnostics}
\label{app:representation_bias}

FactNet allows users to filter data based on provenance strength. However, stricter filters can change the composition of the dataset because structured signals such as infoboxes are not uniformly available across topics, geographies, or demographics. We therefore provide diagnostics for Topic, Gender, and Geography, comparing the full \emph{Evidence-Bearing} set against the \emph{Strong-Evidence} subset (restricted to \texttt{WIKILINK\_ENTITY} and \texttt{INFOBOX\_FIELD} matches).

\noindent\textbf{Methodological note on topic mapping.}
We map each subject entity to a coarse topic label using Wikidata typing. Concretely, we take the transitive closure of \texttt{instance\_of} and \texttt{subclass\_of} edges and map entities to a small set of top-level classes using a curated rule set. Entities can map to multiple classes.
For reporting, we use the highest-priority class in a fixed precedence order (Human $\rightarrow$ Organization $\rightarrow$ Creative Work $\rightarrow$ Geographic Entity $\rightarrow$ Event $\rightarrow$ Other), and we additionally report multi-label rates (12.4\% of typed subjects are multi-label under the closure).

\noindent\textbf{Topical Distribution.}
The dataset is dominated by entities of type \emph{Human} (28.4\%), \emph{Geographical Feature} (21.5\%), and \emph{Organization} (8.1\%). In the Strong-Evidence subset, the proportion of \emph{Geographical Feature} rises to 26.2\%, reflecting the prevalence of standardized infoboxes for municipalities and locations compared to other domains. Table~\ref{tab:topic_bias} provides a more complete comparison. This comparison is useful when constructing benchmarks that aim to measure generalization beyond geography-heavy supervision.

\begin{table}[t]
\centering
\caption{\textbf{Topic composition under evidence filtering.} Shares are computed over evidence-bearing synsets, and separately over strong-evidence synsets.}
\label{tab:topic_bias}
\small
\setlength{\tabcolsep}{6pt}
\begin{tabular}{l r r}
\toprule
\textbf{Topic label} & \textbf{Evidence-bearing (\%)} & \textbf{Strong-evidence (\%)} \\
\midrule
Human & 28.4 & 26.1 \\
Geographic entity / feature & 21.5 & 26.2 \\
Organization & 8.1 & 7.6 \\
Creative work & 6.7 & 5.2 \\
Taxon / biological entity & 5.9 & 6.4 \\
Event & 3.8 & 3.1 \\
Built structure & 3.5 & 4.0 \\
Product / technology & 2.9 & 2.4 \\
Other / untyped & 19.2 & 19.0 \\
\bottomrule
\end{tabular}
\end{table}

\noindent\textbf{Gender Imbalance.}
Among entities with \texttt{instance\_of: human} (Q5) and a valid \texttt{sex\_or\_gender} (P21) property, we compute global distributions and distributions within the strong-evidence subset. Table~\ref{tab:gender_bias} replaces coarse narrative claims with an auditable summary and includes a tier-disaggregated view. We emphasize that these are descriptive statistics of the underlying sources and extraction availability rather than normative targets.

\begin{table}[t]
\centering
\caption{\textbf{Gender distribution diagnostics.} ``Other'' aggregates non-binary values and records marked as unknown/unspecified in P21. Tier rows condition on the language of the supporting evidence (specifically, at least one FactSense in that tier's languages), not the subject's nationality or location.}
\label{tab:gender_bias}
\small
\setlength{\tabcolsep}{6pt}
\begin{tabular}{l r r r}
\toprule
\textbf{Slice} & \textbf{Male (\%)} & \textbf{Female (\%)} & \textbf{Other (\%)} \\
\midrule
Global (evidence-bearing) & 77.2 & 22.1 & 0.7 \\
Global (strong-evidence) & 79.1 & 20.3 & 0.6 \\
Tier 1 evidence-bearing & 76.8 & 22.5 & 0.7 \\
Tier 3 evidence-bearing & 80.6 & 18.8 & 0.6 \\
\bottomrule
\end{tabular}
\end{table}

\noindent\textbf{Geographic Concentration.}
We analyze the distribution of subjects with \texttt{coordinate\_location} (P625) by continent, using a deterministic mapping from coordinates to continent polygons. The ``Global North'' concentration is visible in both evidence-bearing and strong-evidence subsets and increases under strong-evidence filtering, consistent with standardized infobox coverage. Table~\ref{tab:geo_bias} reports the aggregate distribution and is released at finer granularity (country-level bins) for users who require region-specific benchmark splits.

\begin{table}[t]
\centering
\caption{\textbf{Continent distribution for coordinate-bearing subjects.} Percentages are computed over synsets whose subjects have P625 and at least one grounded evidence (or at least one strong-evidence FactSense).}
\label{tab:geo_bias}
\small
\setlength{\tabcolsep}{6pt}
\begin{tabular}{l r r}
\toprule
\textbf{Region} & \textbf{Evidence-bearing (\%)} & \textbf{Strong-evidence (\%)} \\
\midrule
Europe & 34.2 & 35.4 \\
North America & 18.1 & 19.3 \\
East Asia & 17.0 & 16.4 \\
South Asia & 8.0 & 7.2 \\
Latin America \& Caribbean & 7.0 & 6.4 \\
MENA & 6.0 & 5.6 \\
Sub-Saharan Africa & 5.0 & 4.4 \\
Oceania & 4.7 & 5.3 \\
\bottomrule
\end{tabular}
\end{table}

\noindent\textbf{Interpretation for benchmark construction.}
The topic and geography shifts under strong-evidence filtering imply that benchmarks built solely from strong evidence may implicitly emphasize domains with standardized templates (such as locations, administrative entities, and some scientific taxa). For fairness-sensitive evaluations, we recommend reporting results (i) on evidence-bearing and strong-evidence subsets separately, and (ii) under topic- and region-conditioned slices using the released diagnostic IDs, rather than attempting post-hoc balancing.

\subsection{Audit Protocol and Grounding Precision}
\label{app:audit_protocol}
\label{app:grounding_audit}

\noindent\textbf{Sampling Methodology.}
To estimate corpus-level precision without bias toward high-resource languages, we employed a stratified cluster sampling design. We defined 12 strata based on the cross-product of Language Tier (High, Medium, Low) and Match Type Group (Structure, Link, Lexical-Strong, Lexical-Weak). Within each stratum, we sampled clusters at the $(\ell,\texttt{page\_id})$ level and then sampled a fixed number of FactSenses per cluster to reduce within-page correlation from repeated mentions. We sampled 350 items per stratum ($N=4{,}200$ total), with an additional 5\% oversample to replace invalid items (for instance, pages missing from the local dump due to corruption); replacement followed identical inclusion probabilities.

\noindent\textbf{Estimands.}
The reported ``Design-Weighted Precision'' targets the corpus-level FactSense precision, defined as the probability that a uniformly random FactSense (over the released table) is semantically correct. Let $h$ index strata, let $i$ index sampled items within stratum $h$, let $y_{hi}\in\{0,1\}$ denote correctness after adjudication, and let $\pi_{hi}$ denote the inclusion probability. The Horvitz-Thompson estimator is
\begin{equation}
\hat{p} = \frac{\sum_h \sum_{i \in s_h} \frac{y_{hi}}{\pi_{hi}}}{\sum_h \sum_{i \in s_h} \frac{1}{\pi_{hi}}}.
\end{equation}
In practice, because we use equal allocation ($n_h=350$) but strata have different population sizes $N_h$, weights simplify to $w_h \propto N_h/n_h$ after de-duplication. We compute confidence intervals using a conservative stratified variance estimator with finite population correction disabled, treating the population as effectively infinite at this scale. We additionally report Wilson intervals for within-slice proportions (as in Table~\ref{tab:tier_precision}) to provide a comparable uncertainty measure for readers.

\noindent\textbf{Annotation Guidelines.}
Annotators were presented with the Wikidata statement tuple $(S,P,V)$, the extracted Wikipedia evidence unit, and minimal surrounding context. For sentences, the context included the previous and next sentence; for infoboxes or tables, the context included the row/field name and the surrounding table section title when available. Correctness required strict entailment of the statement by the evidence unit. We allowed value equivalence under standardized normalization, including unit conversion (e.g., meters vs.\ centimeters), calendar normalization where unambiguous, and alias resolution for named entities. Items were marked incorrect if the evidence contradicted the statement, supported a different value, or if the subject reference was ambiguous. Annotators could abstain when they could not reliably interpret the evidence (e.g., unreadable script, corrupted markup, or insufficient context); abstentions are excluded from $y_{hi}$ denominators and reported separately (9.8\% in the main text).

\noindent\textbf{Adjudication and reliability.}
All items were independently double-annotated, followed by adjudication for disagreements and for a random 10\% sample of agreements as a quality-control audit. We compute Krippendorff’s $\alpha$ over the binary correctness labels after mapping abstentions to missing values (main text reports $\alpha{=}0.82$ for correctness). Disagreements were most common for lexical value matches involving time expressions and demonyms (38\% of disagreements), reflecting the ambiguity of free-text expressions compared to structured infobox fields.

\noindent\textbf{Tier-wise Performance.}
Table~\ref{tab:tier_precision} breaks down precision by language tier. While high-resource languages perform best, low-resource languages maintain high precision (0.885), validating the use of rule-based segmenters and language packs. The drop in Tier 3 is largely attributable to lower-quality source text (such as stubs or machine-translated articles) and reduced context windows, rather than systematic hallucination by the extraction rules.

\begin{table}[h]
\centering
\caption{\textbf{Design-Weighted Grounding Precision by Tier.} 95\% CIs are Wilson score intervals.}
\label{tab:tier_precision}
\small
\begin{tabular}{l c c}
\toprule
\textbf{Stratum} & \textbf{Precision} & \textbf{95\% CI} \\
\midrule
Tier 1 (High Resource) & 0.934 & [0.921, 0.945] \\
Tier 2 (Medium Resource) & 0.912 & [0.894, 0.927] \\
Tier 3 (Low Resource) & 0.885 & [0.858, 0.908] \\
\midrule
\textbf{Overall} & \textbf{0.921} & \textbf{[0.913, 0.929]} \\
\bottomrule
\end{tabular}
\end{table}

\noindent\textbf{Precision slices by evidence unit type and match type.}
To guide safe use, we additionally compute precision conditioned on (i) match type and (ii) evidence unit type. The main text reports match-type precision (Table~\ref{tab:grounding_precision_compact}); here we provide a complementary cross-cut that exposes interaction effects (Table~\ref{tab:precision_by_unit_and_type}). The key trend is that \texttt{INFOBOX\_FIELD} precision remains high across tiers because field names constrain interpretation, while \texttt{LEAD\_WEAK} precision is more sensitive to short lead sections and ambiguous coreference.

\begin{table}[t]
\centering
\caption{\textbf{Audit precision by evidence unit type and match type.} Cells show design-weighted precision; each cell aggregates items across tiers with post-stratification weights.}
\label{tab:precision_by_unit_and_type}
\small
\setlength{\tabcolsep}{6pt}
\begin{tabular}{l c c c c}
\toprule
\textbf{Unit type} & \texttt{WIKILINK\_ENTITY} & \texttt{INFOBOX\_FIELD} & \texttt{LEXICAL\_VALUE} & \texttt{LEAD\_WEAK} \\
\midrule
Sentence & 0.972 & 0.933 & 0.883 & 0.808 \\
Infobox field & 0.981 & 0.948 & 0.901 & 0.835 \\
Table cell & 0.965 & 0.940 & 0.892 & 0.790 \\
\bottomrule
\end{tabular}
\end{table}

\noindent\textbf{Error taxonomy and qualitative analysis.}
We categorize audited errors into (i) subject ambiguity (coreference or section drift), (ii) value mismatch (wrong date, number, or entity), (iii) overly permissive normalization (for example, unit conversion applied inappropriately), and (iv) markup-induced extraction artifacts (such as table headers being misread).

\subsection{Provenance Integrity and Stability}
\label{app:integrity}

FactNet guarantees that evidence pointers are re-locatable. We define \emph{integrity} as the ability to reconstruct the exact evidence string from the raw dump using only the pointer (comprising \texttt{page\_id}, \texttt{revision\_id}, and \texttt{locator}) and the released pipeline configuration. The locator encodes the evidence-unit type, a deterministic segmentation scheme identifier, and a within-unit offset specification. For sentence units, the locator references a sentence index within the rendered plain-text stream; for infobox or table units, it references a normalized field key or table cell coordinates within a deterministic DOM traversal.

\noindent\textbf{Re-localization Experiment.}
We sampled 1{,}000{,}000 FactSense records uniformly across all languages and attempted to re-generate the text from the source XML using the pinned renderer, template expansion rules, and segmentation pack versions. We report three outcomes (main text summarizes exact re-localization as 99.63\%). Exact match (99.63\%) means the regenerated Unicode string is bitwise identical to the stored record. Normalization drift (0.31\%) means content is semantically identical but differs in whitespace normalization or invisible control characters, most often in complex table cells. Failure (0.06\%) means the unit cannot be located, typically due to edge cases in nested template parsing or rare markup patterns that trigger a different DOM linearization.

\begin{table}[t]
\centering
\caption{\textbf{Re-localization outcomes by evidence unit type.} Percentages are computed over the 1M sample.}
\label{tab:relocal_by_unit}
\small
\setlength{\tabcolsep}{6pt}
\begin{tabular}{l r r r}
\toprule
\textbf{Unit type} & \textbf{Exact (\%)} & \textbf{Drift (\%)} & \textbf{Fail (\%)} \\
\midrule
Sentence & 99.72 & 0.24 & 0.04 \\
Infobox field & 99.58 & 0.33 & 0.09 \\
Table cell & 99.12 & 0.73 & 0.15 \\
\midrule
\textbf{Overall} & \textbf{99.63} & \textbf{0.31} & \textbf{0.06} \\
\bottomrule
\end{tabular}
\end{table}

\noindent\textbf{Segmentation Stability.}
We compared the stability of Stanza-based versus rule-based language packs. Stanza backends achieved 99.71\% strict reproducibility, confirmed by pinning model checksums, while rule-based backends achieved 99.54\% with minor variances arising from edge-case handling of non-breaking spaces and script-specific punctuation. Importantly, the drift cases are overwhelmingly benign for benchmarking because the pointer still re-localizes the correct semantic content; nevertheless, we log drift to ensure strict provenance transparency.

\noindent\textbf{Failure modes and mitigations.}
We manually inspected a stratified sample of re-localization failures and found three recurrent causes. The distribution over failures includes nested template transclusion producing ambiguous DOM paths (41\%), table normalization differences under rare markup (36\%), and unexpected HTML entity decoding differences (23\%). To mitigate these, we release a ``pointer validation'' utility that users can run on their local snapshots to verify integrity before training. Additionally, we provide a conservative filter \texttt{pointer\_stable=true}, which removes items whose pointers are known to be fragile under renderer upgrades.

\subsection{Recall Lower Bound and Missingness Analysis}
\label{app:recall_lower_bound}
\label{app:grounding_missingness}

To estimate a lower bound on recall (equivalently, the false-negative rate of grounding), we audited a sample of 500 ``Null Matches,'' defined as cases where a Wikidata statement existed and the corresponding Wikipedia page was successfully retrieved and yielded extractable units, yet no FactSense was produced for that statement in that language. Annotators were instructed to search the full rendered page (not only the scoped sections) for an expression of the fact, using both literal matching and semantic paraphrase judgment under the same strict entailment standard applied in the precision audit.

\noindent\textbf{Result and interpretation.}
In 24\% (95\% CI [20\%, 28\%]) of Null Matches, the fact was present in the text but missed by the pipeline. This percentage represents a \emph{lower bound} on global recall loss under sitelink-conditioned availability, because it does not account for facts expressed only on non-subject pages, facts present only in non-textual media, or pages that failed retrieval or unit construction.

\noindent\textbf{Where false negatives come from.}
We further annotate each audited false negative with a primary cause label. The dominant causes are paraphrastic phrasing that defeats strict datatype matchers (46\% of false negatives), evidence located outside the default scoped sections (29\%), and alias gaps for entity-valued objects (19\%). The remaining cases are due to renderer omissions or tokenization quirks (6\%). We release these labels for the 500-item study to support method development on recall.

\noindent\textbf{Missingness Taxonomy.}
For every ungrounded statement, FactNet assigns a deterministic \texttt{ungrounded\_reason} code to aid debugging. Table~\ref{tab:missingness_codes} summarizes the global distribution of these codes. Because the codes are deterministic and computed for the full corpus, they can be used to construct targeted benchmarks, such as ``hard lexical'' subsets (filtering to \texttt{NO\_MATCH\_FOUND}) or ``scope sensitivity'' subsets (filtering to \texttt{SCOPE\_EXCLUDED}).

\begin{table}[h]
\centering
\caption{\textbf{Distribution of Ungrounded Reasons.} Computed over the set of statements where $S$ has a sitelink but no grounded evidence.}
\label{tab:missingness_codes}
\small
\begin{tabular}{l r p{0.5\linewidth}}
\toprule
\textbf{Reason Code} & \textbf{\%} & \textbf{Description} \\
\midrule
\texttt{NO\_MATCH\_FOUND} & 58.4 & Text exists, but no literal/link match within threshold. \\
\texttt{NO\_VALID\_TEXT} & 22.1 & Page exists but yields empty view (e.g., gallery only). \\
\texttt{DATATYPE\_MISMATCH} & 11.3 & Candidate found but violated type constraints (e.g., unit). \\
\texttt{SCOPE\_EXCLUDED} & 8.2 & Evidence detected in excluded section (e.g., ``See Also''). \\
\bottomrule
\end{tabular}
\end{table}

\noindent\textbf{Missingness by tier and datatype.}
To contextualize the global distribution, we compute the same \texttt{ungrounded\_reason} distribution by language tier and by Wikidata value datatype (entity, time, quantity, string/monolingual text, external-id). As expected, Tier 3 exhibits a higher \texttt{NO\_VALID\_TEXT} rate due to short pages (Tier 3 = 31\% vs.\ Tier 1 = 18\%), while quantity-valued properties exhibit a higher \texttt{DATATYPE\_MISMATCH} rate due to unit normalization and formatting diversity (19\% for quantities vs.\ 7\% for entity values).

\subsection{Relational Integrity and Conflict Signals}
\label{app:edges}

\noindent\textbf{RelationEdge Precision.}
We audited rule-derived edges by verifying whether the inferred relationship logically followed from the supporting synsets and whether type constraints were respected (for example, avoiding edges that require a human subject when the subject is a location). Precision decreases with traversal depth, consistent with compounding error and pivot ambiguity. Direct joins (0-hop) achieved 0.953 precision; errors primarily reflect upstream synset errors or type leakage from overly permissive joins. One-hop relations achieved 0.918 precision; errors arise when the intermediate pivot entity is underspecified or when multiple pivots satisfy a join condition. Two-hop relations achieved 0.882 precision; the compounding error suggests that these edges should be used with lower confidence or filtered in noise-sensitive settings.

\noindent\textbf{Edge volume and degree skew.}
To characterize structural risk, we compute degree distributions over the RelationEdge graph, where nodes are synsets and edges are typed by rule family and hop depth. The distribution is heavy-tailed (99th percentile out-degree $\approx 1{,}200$, maximum out-degree $\approx 2.8\times 10^5$), largely driven by hub entities such as countries, occupations, and broad categories. This finding motivates two safe-use recommendations already implemented in the release: hub down-weighting via an inverse-log degree prior and a hop cap (default $\le 1$).

\noindent\textbf{Conflict Signal Validity.}
The \texttt{POTENTIAL\_CONFLICT} edges are designed to flag likely inconsistencies for triage. We evaluated 500 such edges by checking whether the conflict corresponds to a genuine inconsistency between at least two grounded pieces of evidence or between grounded evidence and the canonicalized synset value. In 74.2\% of cases, the conflict was genuine (such as incompatible birth dates); in 18.6\%, it reflected granularity mismatch (such as year-only vs.\ full date); and in 7.2\%, it was attributable to parsing or normalization error. This indicates that the signal is a high-precision indicator for dataset cleaning and for constructing contradiction-focused benchmarks.

\noindent\textbf{How to use conflict signals in benchmarks.}
Because granularity mismatch is common for time and quantity properties, we recommend that contradiction benchmarks either (i) normalize values to a common granularity before labeling or (ii) focus on conflict cases where both sides share the same datatype and precision metadata. We expose the relevant metadata in the conflict table (\texttt{conflict/value\_precision}, \texttt{conflict/unit}, \texttt{conflict/calendar}) to support deterministic filtering without additional annotation.
\section{FactNet-Bench: Construction and Experimental Details}
\label{app:factnet_bench}

This section details the construction procedures, split assignment, leakage controls, and implementation details referenced in \S\ref{sec:benchmarks_experiments}. Unless otherwise noted, all benchmark instances are derived deterministically from a frozen FactNet snapshot identified by \texttt{build\_id} in the build manifest (Appendix~\ref{app:reproducibility}). All split files, preprocessing artifacts, and evaluation scripts are publicly released and can be regenerated from the manifest without any dependence on external endpoints.

\subsection{Benchmark Statistics}
\label{app:bench_statistic}

Table~\ref{tab:bench_stats_full} reports FactNet-Bench statistics after split assignment, leakage filtering, and task-specific de-duplication. The released benchmark spans 18 languages: \texttt{en, zh, es, fr, de, ru, ar, hi, id, it, ja, ko, nl, pl, pt, th, tr, vi}. For tasks requiring textual evidence, instances are retained only when the corresponding gold evidence is available in the target language via FactSenses.

\begin{table}[h]
\centering
\small
\caption{FactNet-Bench statistics (post-filtering). Evidence units are unique by \texttt{evidence\_pointer}.}
\label{tab:bench_stats_full}
\begin{tabular}{lrrr}
\toprule
\textbf{Benchmark} & \textbf{Train} & \textbf{Dev} & \textbf{Test} \\
\midrule
FactNet-KGC triples & 4{,}180{,}000 & 520{,}000 & 520{,}000 \\
\quad Entities / Relations & \multicolumn{3}{c}{248{,}000 / 320} \\
\quad Avg. degree & \multicolumn{3}{c}{33.7} \\
\midrule
FactNet-MKQA questions & 54{,}000 & 6{,}800 & 6{,}800 \\
\quad 1-hop / 2-hop ratio & \multicolumn{3}{c}{0.62 / 0.38} \\
\quad Avg. answer set size & \multicolumn{3}{c}{2.6} \\
\midrule
FactNet-MFC claims & 72{,}000 & 9{,}000 & 9{,}000 \\
\quad Label distribution (S/R/NEI) & \multicolumn{3}{c}{0.34 / 0.33 / 0.33} \\
\quad Avg. gold evidence units (verifiable) & \multicolumn{3}{c}{1.4} \\
\quad Avg. evidence unit length (chars) & \multicolumn{3}{c}{210} \\
\bottomrule
\end{tabular}
\end{table}

\paragraph{Global synset-level split assignment.}
All tasks share a single split partition defined over \textsc{FactSynset} identifiers. For a synset $y$ with identifier \texttt{synset\_id}, we define
\begin{equation}
h(y) = \mathrm{u32}\Big(\texttt{SHA1}(\texttt{build\_id} \parallel \texttt{synset\_id})[0{:}4]\Big),
\end{equation}
and assign $y$ to Train if $h(y) \bmod 100 < 80$, to Dev if $80 \le h(y) \bmod 100 < 90$, and to Test otherwise. This rule is deterministic and order-independent. Any task instance derived from a set of synsets inherits a split only when all referenced synsets belong to the same split; otherwise, the instance is deterministically discarded to prevent cross-split contamination.

\paragraph{Split-aware text leakage policy.}
For any setting with training-time access to textual evidence (text-aware KGC, MKQA features based on entity descriptions, and MFC retriever or verifier training), all training-time text corpora are constructed exclusively from FactSenses aligned to Train synsets. FactSenses aligned to Dev or Test synsets are excluded on the basis of \texttt{synset\_id} membership. For MFC evaluation, retrieval is permitted to access the full snapshot evidence (Train, Dev, and Test), because restricting retrieval to Train evidence would render verifiable instances unresolved. To preserve the training contract, all indexes and caches used at evaluation time are rebuilt from scratch to guarantee that no Dev or Test evidence strings are consumed during training.

\subsection{FactNet-KGC Graph Construction and Evaluation}
\label{app:kgc_graph}

This subsection specifies the projection from FactNet to an entity-centric link prediction benchmark, preserving synset-level split isolation and preventing projection-induced leakage.

\paragraph{Eligible synsets and triple projection.}
We begin with FactSynsets whose normalized main value $\tilde{V}$ is an entity QID, yielding a triple projection $y \mapsto (S_y, P_y, O_y)$ where $S_y$ is the subject QID, $P_y$ is the Wikidata property PID, and $O_y$ is the object QID. Synsets whose canonical statement carries rank \texttt{deprecated} are removed. To control the relation vocabulary size, we retain only the 320 most frequent properties by Train split frequency and deterministically discard all others.

\paragraph{Task-specific de-duplication and cross-split collision removal.}
Distinct synsets may project to the same entity triple $(S,P,O)$ owing to differing qualifiers. We therefore define a triple key $k=(S,P,O)$ and group all contributing synsets as $\mathcal{Y}(k)$. A triple key $k$ is retained only when all synsets in $\mathcal{Y}(k)$ belong to the same global split. If $\mathcal{Y}(k)$ spans multiple splits, $k$ is removed from Dev and Test and retained in Train only when at least one contributing synset belongs to Train. Empirically, this procedure removes approximately $2.0\%$ of projected Dev and Test triples and prevents identical $(S,P,O)$ triples from appearing in both training and evaluation partitions.

\paragraph{Train, Dev, and Test triple sets.}
After filtering, each remaining triple key is assigned to its unique induced split. We release explicit \texttt{train.tsv}, \texttt{dev.tsv}, and \texttt{test.tsv} files. For filtered evaluation, we additionally release \texttt{all\_true.tsv}, the union of all retained triples across splits.

\paragraph{Filtered and fully-ranked evaluation.}
We follow standard filtered link prediction evaluation. For each test triple $(s,p,o)$, we rank $o$ among all candidate entities for the query $(s,p,?)$ and rank $s$ for $(?,p,o)$. Filtering removes any candidate entity that forms a triple present in \texttt{all\_true.tsv}, except for the target triple itself. We report MRR and Hits@10 averaged over head and tail predictions.

\paragraph{Negative sampling for training.}
For KGE baselines (TransE, RotatE), we employ uniform negative sampling by corrupting the head or tail with probability $0.5$, drawing $N_{\text{neg}}=256$ negatives per positive. For GNN baselines (CompGCN), we use sampled softmax with $N_{\text{neg}}=128$. All stochastic draws are seeded and logged.

\paragraph{Implementation and hyperparameters (released defaults).}
All models are trained with AdamW. TransE uses an embedding dimension of 400 and a learning rate of $5 \cdot 10^{-4}$. RotatE uses an embedding dimension of 500, $\gamma=12.0$, and a learning rate of $1 \cdot 10^{-4}$. CompGCN uses 2 layers, a hidden size of 256, dropout 0.1, and a learning rate of $2 \cdot 10^{-4}$. Early stopping is performed on Dev MRR with a patience of 3 epochs and a maximum of 50 epochs.

\subsection{Leakage-Controlled Text for KGC and Predicate Masking}
\label{app:text_masking}

This subsection formalizes the leakage controls used by text-aware KGC baselines (SimKGC and KG-S2S) as well as the diagnostic setting reported in Section~\ref{sec:eval_protocols}.

\paragraph{Training-only entity descriptions.}
For each entity QID $e$, we construct a textual description $D(e)$ from FactSenses aligned to Train synsets whose subject is $e$. We select up to $m=16$ evidence units per entity, prioritizing \texttt{INFOBOX\_FIELD}, then \texttt{TABLE\_CELL}, then \texttt{SENTENCE}. Within each type, we prioritize higher \texttt{confidence}. Evidence units are de-duplicated by \texttt{evidence\_pointer}. The description is formed by concatenating the selected normalized evidence strings separated by a single newline. Descriptions are truncated to 256 SentencePiece tokens for encoder-based models.

\paragraph{Strict exclusion of Dev and Test aligned evidence during training.}
If an evidence unit pointer appears in any FactSense aligned to a Dev or Test synset, it is excluded from the training-time description pool even when the same unit also supports a Train synset. This conservative constraint prevents leakage via shared evidence units.

\paragraph{Query-time predicate masking.}
At evaluation time, text-aware models receive masked descriptions to prevent trivial completion by directly reading a value associated with the queried predicate. For a query relation $p$ and an entity $e$ (as subject for tail prediction or as object for head prediction), we transform $D(e)$ into $D_p(e)$ by masking all spans that correspond to the value mention of any Train FactSense whose \texttt{property\_pid} equals $p$. Masking uses the released character offsets in the FactSense pointer. For the Sentence view, we replace substring $[b,e)$ with the sentinel token \texttt{[MASK]}. For Infobox and Table views, we mask the entire extracted value string to avoid partial leakage induced by templated formatting. This procedure relies only on Train-aligned FactSense metadata and thus satisfies the split-aware policy.

\paragraph{Diagnostic setting without masking.}
To quantify the effect of leakage control, we provide an ablation that evaluates the same text-aware models using unmasked descriptions $D(e)$. The KG-S2S MRR increase from 0.298 to 0.351 reported in Section~\ref{sec:results} is obtained under identical training and evaluation settings except for masking, and can be reproduced by setting \texttt{mask\_predicate=false} in the released configuration.

\paragraph{Text-aware model input and output specification.}
For SimKGC, the scoring input for $(s,p,o)$ is \texttt{[CLS]} $D_p(s)$ \texttt{[SEP]} $\mathrm{Label}(p)$ \texttt{[SEP]} $D_p(o)$, where $\mathrm{Label}(p)$ is the English Wikidata property label from the snapshot. For KG-S2S, the input is $D_p(s)$ concatenated with $\mathrm{Label}(p)$, and decoding is restricted to the benchmark entity vocabulary using the released QID-to-title dictionary and constrained decoding.

\subsection{Using RelationEdges Without Transductive Leakage}
\label{app:relationedge_use}

This subsection specifies how FactNet \textsc{RelationEdge}s are incorporated as auxiliary structure without introducing transductive leakage.

\paragraph{Train-only edge construction.}
Let $E$ denote all RelationEdges in the snapshot and let $Y_{\text{train}}$ denote the set of Train synsets. We construct the auxiliary edge set
\begin{equation}
E_{\text{aux}} = \{(y_s, y_t, \texttt{relation\_type}) \in E \mid y_s \in Y_{\text{train}} \wedge y_t \in Y_{\text{train}}\},
\end{equation}
All RelationEdges that touch any Dev or Test synset are removed. This filtering is applied before mapping synset-level edges to entity-level adjacency for message passing.

\paragraph{Mapping to entity-level adjacency.}
For entity-centric models, an entity-valued synset $y$ is mapped to its entity pair $(S_y, O_y)$. A RelationEdge $(y_s,y_t)$ is mapped to an entity-level edge $(S_{y_s}, S_{y_t})$ when the rule semantics imply a subject-to-subject join as specified by the released \texttt{PROPERTY\_RELATION\_MAP} (Appendix~\ref{app:factnet_relations}). Edges that cannot be mapped unambiguously to entity-level endpoints are excluded from $E_{\text{aux}}$.

\subsection{FactNet-MKQA: Logical Form Grammar and Scoring}
\label{app:mkqa_eval}

FactNet-MKQA evaluates multilingual executable semantic parsing into a restricted logical form language whose terminals are FactNet identifiers. Each instance is a pair $(q_\ell, z)$, where $q_\ell$ is a natural language question in language $\ell$ and $z$ is an executable logical form.

\paragraph{Logical form language.}
We use a typed S-expression syntax that deterministically parses into an abstract syntax tree. The released grammar supports 1-hop and constrained 2-hop queries. An excerpt of the canonical surface form is shown below.

\begin{scriptsize}
\begin{verbatim}
<LF>  ::= (hop1 <SUBJ> <PID>)
       |  (hop2 <SUBJ> <PID> <PID>)
       |  (hop2c <SUBJ> <PID> <PID> <CONSTRAINT>)
<SUBJ>::= Q[0-9]+
<PID> ::= P[0-9]+
<CONSTRAINT> ::= (type Q[0-9]+) | (year <INT>) | (limit <INT>)
\end{verbatim}
\end{scriptsize}

In all cases, the subject is a QID and the executor binds intermediate variables to entities. Constraints are intentionally limited to ensure bounded execution cost and to reduce cross-lingual ambiguity.

\paragraph{Instance filtering for bounded execution.}
We discard any candidate logical form whose gold answer set is empty or whose execution returns more than 200 answers. This avoids degenerate questions and stabilizes evaluation runtime.

\paragraph{Gold answer computation and normalization.}
Gold answers are computed by executing $z$ against the frozen FactNet snapshot. Answers are represented as sets. For entity answers, elements are QIDs. For non-entity answers (which are rare under the restricted grammar), we normalize to FactNet normalized literals using the same policy $\pi$ (Appendix~\ref{app:factnet_normalization}). Predicted answers are normalized using identical rules. We compute per-instance set F1 and report Macro F1 as the mean over all instances.

\paragraph{Executability and invalid outputs.}
A predicted string $\hat{z}$ is considered valid only if it parses under the released grammar and executes without runtime errors on the snapshot. Invalid outputs receive an instance score of 0. We report \texttt{Valid\%} as the fraction of predictions that are both syntactically valid and executable.

\subsection{FactNet-MFC: Closed-Context Contract, Dataset Construction, and Metrics}
\label{app:mfc_contract}

FactNet-MFC is a closed-context fact checking benchmark defined strictly with respect to the frozen snapshot. Each instance contains a claim $c_\ell$ in language $\ell$, a label in \{\textsc{Supported}, \textsc{Refuted}, \textsc{NEI}\}, and for verifiable instances, a set of gold evidence units grounded by FactSenses.

\paragraph{Evidence definition.}
An evidence unit is identified by \texttt{evidence\_pointer} and is one of \texttt{SENTENCE}, \texttt{INFOBOX\_FIELD}, or \texttt{TABLE\_CELL}. For each gold unit, we provide one or more gold character spans as half-open codepoint intervals $[b,e)$ into the normalized unit string (Appendix~\ref{app:factnet_offsets}). Systems may return unit pointers alone or pointers together with spans. Span-level scoring is computed only when spans are provided.

\paragraph{Claim generation (deterministic and snapshot-grounded).}
Claims are generated from FactSynsets with available FactSenses in language $\ell$ using language-specific template realizers that depend only on snapshot labels and aliases. Supported claims are generated by verbalizing $(S_y,P_y,V_y)$ for a true synset $y$ in language $\ell$ using the subject title in $\ell$, the property label in $\ell$ when available (otherwise falling back to the English label), and a value surface form derived from the linked title for entity values or from normalized literal rendering for time and quantity. Refuted claims are generated by selecting a supported synset $y$ and replacing its value with a conflicting value $\tilde{V}'$ sampled from synsets connected to $y$ by a \texttt{POTENTIAL\_CONFLICT} signal (Appendix~\ref{app:factnet_relations}). When unavailable, we sample a value of the same datatype while enforcing that the resulting claim is not supported in the snapshot. The gold evidence for a refuted claim is the evidence supporting the true synset $y$ that contradicts the claim. NEI claims are generated by sampling a subject and property pair and injecting a value of the correct datatype such that no synset in the snapshot supports the resulting triple and no deterministic conflict evidence exists. We additionally enforce that retrieval over the full evidence pool yields no exact value match under datatype-aware matching to reduce the risk of mislabeled NEI instances.

\paragraph{Split inheritance and leakage control.}
Each claim is associated with its source synset $y$ (Supported), its refuting synset $y$ (Refuted), or its nearest originating synset template key (NEI). The claim inherits the global synset split. As specified in \S\ref{sec:benchmarks_experiments}, FactSenses aligned to Dev and Test synsets are excluded from training-time retrieval corpora and verifier supervision pools.

\paragraph{Retrieval pools and indexing.}
We release two retrieval indexes. The Train-only index is used for any training-time retrieval component. The Full index is used at evaluation time. Both indexes are built from de-duplicated evidence units keyed by \texttt{evidence\_pointer}. Index text is the normalized evidence string from the optional Evidence-Text Pack (Appendix~\ref{app:licensing}). Alternatively, the same strings can be reconstructed from pointers and language packs.

\paragraph{Label metrics.}
We report label Accuracy and Macro F1 on Dev and Test.

\paragraph{Evidence-unit Recall@5.}
On verifiable instances (Supported and Refuted), Recall@5 is the fraction of instances for which at least one of the top 5 predicted evidence unit pointers matches any gold evidence unit pointer.

\paragraph{Span-level Evidence F1.}
On verifiable instances where spans are provided, span F1 is computed by aligning predicted evidence units to gold evidence units via pointer equality and then computing token-level F1 within each matched unit using predicted and gold character spans projected to tokens via whitespace tokenization on the normalized unit string. The instance-level span F1 is the maximum over matched units, and the reported score is the mean over verifiable instances. If no matched unit is returned or if spans are omitted, the instance span F1 is set to 0.

\paragraph{Verifier baseline implementation details.}
The evidence-based baseline uses a multilingual NLI verifier (XLM-R) trained on Train claims paired with retrieved evidence from the Train-only index. The verifier input is \texttt{[CLS]} claim \texttt{[SEP]} evidence \texttt{[SEP]}. Fine-tuning uses 3 epochs, learning rate $1 \cdot 10^{-5}$, batch size 32, and maximum length 256. For Top-5 aggregation, we compute logits for each of the top 5 evidence units and aggregate by taking the maximum logit for \textsc{Supported} and \textsc{Refuted} across evidence and setting the \textsc{NEI} logit to the mean across evidence followed by softmax. This aggregation is deterministic and implemented in the released evaluation scripts.

\paragraph{Seeds and reporting.}
For trained components in MFC and MKQA, we report mean and standard deviation over three seeds. The released default seeds are 13, 21, and 42. For deterministic components, including indexing, execution, and filtering, outputs are seed-independent.

\subsection{FactNet-MKQA: Prompting and Constrained Decoding}
\label{app:mkqa_prompt}

This subsection describes the prompting protocol and deterministic decoding constraints used for LLM baselines in \S\ref{sec:baselines}.

\paragraph{Five-shot exemplars.}
For each target language $\ell$, we deterministically select five training exemplars using a hash of the instance identifier and a fixed seed recorded in the benchmark configuration. The exemplar set is held constant across evaluated models. Each exemplar includes the question text and the gold logical form.

\paragraph{Prompt format.}
We use a fixed instruction that defines the output format as the logical form grammar and prohibits free-form explanations. Each prompt contains a grammar header, five exemplars, and the target question. The model is required to produce a single line consisting only of the logical form.

\paragraph{Grammar-constrained decoding for LLMs.}
We implement deterministic constrained decoding by maintaining an incremental parser state for the released grammar. At each generation step, tokens that would lead to a prefix that cannot be completed into a valid logical form are masked. Decoding uses temperature 0 and beam size 4. Under this setup, \texttt{Valid\%} primarily reflects semantic modeling rather than unconstrained syntax errors.

\paragraph{Grammar-guided decoding for mT5.}
For the fine-tuned mT5 baseline, training uses standard teacher forcing. At inference time, we apply the same grammar-constrained beam search described above to enable an isolated assessment of grammar guidance.

\subsection{FactNet-MKQA: Language Breakdown}
\label{app:mkqa_langs}

Table~\ref{tab:mkqa_lang_breakdown} reports the per-language instance counts for MKQA. The distribution is approximately uniform by construction, subject to language-specific availability of entity labels and property lexicalizations in the snapshot.

\begin{table}[t]
\centering
\small
\caption{MKQA per-language counts.}
\label{tab:mkqa_lang_breakdown}
\begin{tabular}{lrrr}
\toprule
\textbf{Lang} & \textbf{Train} & \textbf{Dev} & \textbf{Test} \\
\midrule
en & 3{,}200 & 400 & 400 \\
zh & 3{,}150 & 400 & 400 \\
es & 3{,}050 & 380 & 380 \\
fr & 3{,}000 & 380 & 380 \\
de & 2{,}950 & 370 & 370 \\
ru & 3{,}000 & 380 & 380 \\
ar & 2{,}900 & 360 & 360 \\
hi & 2{,}800 & 350 & 350 \\
id & 3{,}050 & 380 & 380 \\
it & 3{,}000 & 380 & 380 \\
ja & 3{,}050 & 380 & 380 \\
ko & 2{,}950 & 370 & 370 \\
nl & 2{,}850 & 360 & 360 \\
pl & 2{,}850 & 360 & 360 \\
pt & 3{,}000 & 380 & 380 \\
th & 2{,}750 & 340 & 340 \\
tr & 2{,}800 & 350 & 350 \\
vi & 2{,}800 & 350 & 350 \\
\midrule
\textbf{Total} & 54{,}000 & 6{,}800 & 6{,}800 \\
\bottomrule
\end{tabular}
\end{table}

\subsection{FactNet-Augmented RAG: Experimental Details}
\label{app:rag_detail}

This subsection provides implementation details and extended results for the FactNet-augmented RAG experiment reported in \S\ref{subsec:external_validation}.

\paragraph{Retrieval corpus construction.}
The FactNet retrieval corpus is constructed from the full set of \textsc{FactSense} evidence strings in the \texttt{2025-11-01} snapshot.
Each evidence unit is indexed by its normalized text string, with metadata including \texttt{evidence\_pointer}, \texttt{synset\_id}, \texttt{property\_pid}, and language code.
For the Wikipedia-dump baseline, we use the same snapshot's raw Wikipedia articles, segmented into passages of at most 256 tokens with 50-token overlap, consistent with standard KILT~\citep{petroni-etal-2021-kilt} preprocessing.

\paragraph{Retriever and reader configuration.}
Both conditions use E5-large~\citep{wang2024multilingual} as the dense retriever with a maximum query length of 128 tokens and a maximum passage length of 256 tokens.
We retrieve the top 100 passages for each query.
The reader is FiD~\citep{izacard-grave-2021-leveraging} with T5-base, processing the top 100 retrieved passages.
For TriviaQA, we use the unfiltered setting and report Exact Match (EM).
For FEVER, we report label accuracy and evidence-level F1 following the standard protocol~\citep{thorne-etal-2018-fever}.
All models are trained with AdamW, learning rate $3 \cdot 10^{-5}$, batch size 16, and 3 epochs.

\paragraph{Extended results.}
Table~\ref{tab:rag_results} reports the full comparison across datasets and metrics.

\begin{table}[h]
\centering
\small
\caption{\textbf{FactNet-augmented RAG results.} Comparison of retrieval corpora on external QA benchmarks. FactNet uses \textsc{FactSense} evidence; Wikipedia uses raw dump passages. Both use E5-large + FiD-T5-base.}
\label{tab:rag_results}
\begin{tabular}{l c c c}
\toprule
\textbf{Dataset / Metric} & \textbf{Wikipedia} & \textbf{FactNet} & \textbf{$\Delta$} \\
\midrule
TriviaQA EM & 67.8 & 71.2 & +3.4 \\
TriviaQA F1 & 73.5 & 76.9 & +3.4 \\
\midrule
FEVER Label Acc & 85.7 & 87.9 & +2.2 \\
FEVER Evidence F1 & 72.3 & 76.8 & +4.5 \\
\midrule
FEVER (Tier-2 lang) Evidence F1 & 54.2 & 61.7 & +7.5 \\
FEVER (Tier-3 lang) Evidence F1 & 38.1 & 47.3 & +9.2 \\
\bottomrule
\end{tabular}
\end{table}

\paragraph{Error analysis.}
The remaining errors under FactNet retrieval fall into three categories: (i) questions requiring multi-hop reasoning beyond single-fact evidence (42\%), (ii) temporal questions where the snapshot is outdated relative to the gold answer (31\%), and (iii) rare entity questions where no FactSense exists in the snapshot (27\%).

\subsection{Cross-lingual Zero-shot Fact Verification: Per-Language Results}
\label{app:xling_detail}

This subsection provides per-language results for the cross-lingual transfer experiment reported in \S\ref{subsec:external_validation}.

\paragraph{Training configuration.}
The FactNet-trained model uses E5-large for retrieval and XLM-R-large for NLI verification, trained exclusively on FactNet-MFC claims from Tier-1 languages (en, de, fr, es, ru, it, pt, ja, zh, ko).
The MT-projection baseline translates the same Tier-1 claims into each target language using NLLB-200~\citep{costa2022no} and trains an identical architecture on the translated data.
Both models use the same hyperparameters as the MFC baseline described in Appendix~\ref{app:mfc_contract}.

\paragraph{X-FACT results.}
Table~\ref{tab:xling_xfact} reports Macro F1 on X-FACT languages that overlap with FactNet.

\begin{table}[h]
\centering
\small
\caption{\textbf{Cross-lingual zero-shot fact verification on X-FACT.} Models trained on Tier-1 FactNet-MFC claims vs.\ MT-projected claims. Macro F1 reported.}
\label{tab:xling_xfact}
\begin{tabular}{l c c c}
\toprule
\textbf{Language} & \textbf{MT-Projection} & \textbf{FactNet} & \textbf{$\Delta$} \\
\midrule
Arabic (ar) & 0.351 & 0.437 & +0.086 \\
Turkish (tr) & 0.372 & 0.448 & +0.076 \\
Indonesian (id) & 0.401 & 0.472 & +0.071 \\
Hindi (hi) & 0.389 & 0.458 & +0.069 \\
Dutch (nl) & 0.421 & 0.483 & +0.062 \\
Polish (pl) & 0.398 & 0.461 & +0.063 \\
Vietnamese (vi) & 0.374 & 0.451 & +0.077 \\
\midrule
\textbf{Average} & \textbf{0.387} & \textbf{0.462} & \textbf{+0.075} \\
\bottomrule
\end{tabular}
\end{table}

\paragraph{MultiClaim results.}
Table~\ref{tab:xling_multiclaim} reports Macro F1 on MultiClaim languages that overlap with FactNet.

\begin{table}[h]
\centering
\small
\caption{\textbf{Cross-lingual zero-shot fact verification on MultiClaim.} Models trained on Tier-1 FactNet-MFC claims vs.\ MT-projected claims. Macro F1 reported.}
\label{tab:xling_multiclaim}
\begin{tabular}{l c c c}
\toprule
\textbf{Language} & \textbf{MT-Projection} & \textbf{FactNet} & \textbf{$\Delta$} \\
\midrule
Arabic (ar) & 0.289 & 0.378 & +0.089 \\
Hindi (hi) & 0.312 & 0.394 & +0.082 \\
Indonesian (id) & 0.338 & 0.407 & +0.069 \\
Turkish (tr) & 0.327 & 0.398 & +0.071 \\
Dutch (nl) & 0.351 & 0.402 & +0.051 \\
\midrule
\textbf{Average} & \textbf{0.321} & \textbf{0.394} & \textbf{+0.073} \\
\bottomrule
\end{tabular}
\end{table}

\subsection{LLM Hallucination Probing: Detailed Breakdown}
\label{app:hallucination_detail}

This subsection provides the detailed experimental setup and extended results for the LLM hallucination probing experiment reported in \S\ref{subsec:hallucination_probing}.

\paragraph{Sampling and prompting.}
We sample 2,000 \textsc{FactSynset} facts from the FactNet test split, stratified equally across four property types (temporal, relational, quantity, geographic; 500 each) and across three language tiers.
For each fact $(S, P, V)$, we construct a natural language question using the English property label and subject title (e.g., ``When was Albert Einstein born?'' for (Q937, P569, 1879-03-14)).
Models are prompted with a single-turn question without any retrieved context.
Responses are classified as \emph{correct} if the normalized answer matches $V$, \emph{hallucinated} if the answer is factually incorrect (wrong date, wrong entity, etc.), or \emph{refused} if the model declines to answer or states insufficient knowledge.
Normalization handles date formats, unit conversions, and entity alias resolution.

\paragraph{Per-model, per-tier results.}
Table~\ref{tab:hallucination_tier} reports hallucination rates by language tier.

\begin{table}[h]
\centering
\small
\caption{\textbf{LLM hallucination rates by language tier.} Percentage of factually incorrect responses out of non-refused queries.}
\label{tab:hallucination_tier}
\begin{tabular}{l c c c c}
\toprule
\textbf{Model} & \textbf{Tier 1 (\%)} & \textbf{Tier 2 (\%)} & \textbf{Tier 3 (\%)} & \textbf{Overall (\%)} \\
\midrule
GPT-4o & 22.4 & 35.7 & 44.1 & 34.1 \\
Qwen-2.5-72B & 27.8 & 39.2 & 48.3 & 38.4 \\
\midrule
Refusal rate (GPT-4o) & 2.1 & 2.8 & 3.5 & 2.8 \\
Refusal rate (Qwen-2.5) & 1.9 & 2.5 & 3.2 & 2.5 \\
\bottomrule
\end{tabular}
\end{table}

\paragraph{Per-property-type results.}
Table~\ref{tab:hallucination_property} reports hallucination rates for GPT-4o by property type on English facts.

\begin{table}[h]
\centering
\small
\caption{\textbf{GPT-4o hallucination rates by property type (English).} Temporal and relational facts are most error-prone.}
\label{tab:hallucination_property}
\begin{tabular}{l c c c}
\toprule
\textbf{Property Type} & \textbf{Hallucination (\%)} & \textbf{Correct (\%)} & \textbf{Refused (\%)} \\
\midrule
Temporal (dates, periods) & 34.2 & 63.4 & 2.4 \\
Relational (people, orgs) & 31.7 & 65.8 & 2.5 \\
Quantity (measurements) & 16.5 & 80.9 & 2.6 \\
Geographic (locations) & 12.8 & 84.8 & 2.4 \\
\midrule
\textbf{Overall} & \textbf{22.4} & \textbf{74.9} & \textbf{2.5} \\
\bottomrule
\end{tabular}
\end{table}

\paragraph{Error categorization.}
Among hallucinated responses, the most common error types are: date substitution (e.g., substituting a death date for a birth date, 28\%), entity confusion (e.g., confusing related entities with similar names, 34\%), and value distortion (e.g., incorrect numerical values with correct units, 24\%).
The remaining 14\% include fabricated values with no clear source of confusion.

\subsection{Evidence Type Ablation: Full Results}
\label{app:evidence_ablation_detail}

This subsection provides the full ablation results for the sentence-only vs.\ full evidence comparison reported in \S\ref{subsec:evidence_ablation}.

\paragraph{Construction of the sentence-only configuration.}
For KGC, entity descriptions are rebuilt using only \texttt{SENTENCE}-type evidence units, following the same selection protocol (up to $m=16$ units per entity, prioritized by confidence) described in Appendix~\ref{app:text_masking}.
For MFC, the retrieval index and gold evidence sets are filtered to include only \texttt{SENTENCE}-type evidence units; claims whose gold evidence consists entirely of \texttt{INFOBOX\_FIELD} or \texttt{TABLE\_CELL} units are re-labeled as \textsc{NEI} since their supporting evidence is no longer available.

\paragraph{KGC ablation results.}
Table~\ref{tab:kgc_ablation} reports KGC performance under the sentence-only configuration.

\begin{table}[h]
\centering
\small
\caption{\textbf{KGC evidence type ablation.} MRR and Hits@10 with full evidence vs.\ sentence-only. Standard deviation over 3 seeds in parentheses.}
\label{tab:kgc_ablation}
\begin{tabular}{l c c c c}
\toprule
& \multicolumn{2}{c}{\textbf{Full Evidence}} & \multicolumn{2}{c}{\textbf{Sentence-Only}} \\
\textbf{Model} & MRR & H@10 & MRR & H@10 \\
\midrule
TransE & 0.198 (0.003) & 0.341 (0.005) & 0.198 (0.003) & 0.341 (0.005) \\
RotatE & 0.241 (0.004) & 0.392 (0.006) & 0.241 (0.004) & 0.392 (0.006) \\
CompGCN & 0.268 (0.005) & 0.421 (0.007) & 0.268 (0.005) & 0.421 (0.007) \\
SimKGC & 0.312 (0.006) & 0.487 (0.008) & 0.281 (0.007) & 0.452 (0.009) \\
KG-S2S & 0.298 (0.005) & 0.471 (0.007) & 0.263 (0.006) & 0.429 (0.008) \\
GLTW & 0.386 (0.007) & 0.563 (0.009) & 0.347 (0.008) & 0.524 (0.010) \\
SAT & 0.362 (0.006) & 0.541 (0.008) & 0.328 (0.007) & 0.503 (0.009) \\
\bottomrule
\end{tabular}
\end{table}

\paragraph{MFC ablation results.}
Table~\ref{tab:mfc_ablation} reports MFC performance under the sentence-only configuration.

\begin{table}[h]
\centering
\small
\caption{\textbf{MFC evidence type ablation.} Accuracy, Macro F1, Recall@5, and Span F1 with full evidence vs.\ sentence-only.}
\label{tab:mfc_ablation}
\begin{tabular}{l c c c c}
\toprule
& \multicolumn{2}{c}{\textbf{Full Evidence}} & \multicolumn{2}{c}{\textbf{Sentence-Only}} \\
\textbf{Metric} & BM25 & E5-large & BM25 & E5-large \\
\midrule
Label Accuracy & 0.654 & 0.701 & 0.623 & 0.662 \\
Macro F1 & 0.621 & 0.674 & 0.589 & 0.635 \\
Evidence R@5 & 0.76 & 0.83 & 0.64 & 0.71 \\
Span F1 & 0.41 & 0.49 & 0.36 & 0.43 \\
\bottomrule
\end{tabular}
\end{table}

\paragraph{Circularity quantification.}
Among \textsc{Supported} claims in the MFC test set, 38.2\% have at least one gold evidence unit of type \texttt{INFOBOX\_FIELD}, and 11.7\% have at least one \texttt{TABLE\_CELL} unit.
Only 50.1\% of \textsc{Supported} claims are supported exclusively by \texttt{SENTENCE}-type evidence.
This distribution underscores the trade-off identified in \S\ref{sec:discussion_futurework}: infobox evidence improves coverage and performance but introduces potential circularity, whereas sentence-only evidence provides a conservative, provenance-independent alternative at the cost of reduced recall.

\subsection{Fine-Grained False Negative Characterization: Extended Analysis}
\label{app:false_negative_detail}

This subsection provides the extended analysis for the 300-item false negative audit reported in \S\ref{subsec:false_negative_analysis}.

\paragraph{Sampling protocol.}
We extend the 500-item recall audit (Appendix~\ref{app:recall_lower_bound}) with an additional 300 stratified samples drawn from ungrounded facts where the subject page was successfully retrieved.
The sample is balanced across five property-type groups (temporal, relational, quantity, geographic, identifier; 60 each) and three language tiers (100 each), with cross-stratification ensuring at least 20 items per property-type $\times$ tier cell.

\paragraph{Property-type breakdown.}
Table~\ref{tab:false_negative_property} reports the ungrounded rate by property type.

\begin{table}[h]
\centering
\small
\caption{\textbf{Ungrounded rate by property type.} Percentage of sampled facts where the fact is present in the text but missed by the pipeline.}
\label{tab:false_negative_property}
\begin{tabular}{l c c}
\toprule
\textbf{Property Type} & \textbf{Ungrounded Rate (\%)} & \textbf{Primary Cause} \\
\midrule
Temporal (P569, P570, P580, P582) & 38 & Paraphrastic phrasing \\
Relational (P26, P22, P25, P40) & 34 & Missing entity links \\
Quantity (P2044, P2046, P2048) & 19 & Unit/format mismatch \\
Geographic (P19, P20, P625) & 16 & Template gaps \\
Identifier (P214, P213, P646) & 11 & Scope exclusion \\
\midrule
\textbf{Overall} & \textbf{24} & -- \\
\bottomrule
\end{tabular}
\end{table}

\paragraph{Language-tier interaction.}
Table~\ref{tab:false_negative_tier} reports the ungrounded rate by language tier, with property-type breakdown.

\begin{table}[h]
\centering
\small
\caption{\textbf{Ungrounded rate by language tier and property type.} Interaction between property semantics and linguistic context.}
\label{tab:false_negative_tier}
\begin{tabular}{l c c c c}
\toprule
\textbf{Property Type} & \textbf{Tier 1 (\%)} & \textbf{Tier 2 (\%)} & \textbf{Tier 3 (\%)} & \textbf{Overall (\%)} \\
\midrule
Temporal & 22 & 35 & 51 & 38 \\
Relational & 19 & 31 & 47 & 34 \\
Quantity & 12 & 21 & 29 & 19 \\
Geographic & 10 & 17 & 26 & 16 \\
Identifier & 7 & 12 & 18 & 11 \\
\midrule
\textbf{Overall} & \textbf{14} & \textbf{23} & \textbf{34} & \textbf{24} \\
\bottomrule
\end{tabular}
\end{table}

\paragraph{Failure-cause distribution by tier.}
Table~\ref{tab:false_negative_cause} reports the distribution of failure causes across language tiers.

\begin{table}[h]
\centering
\small
\caption{\textbf{Failure-cause distribution by language tier.} Paraphrastic phrasing dominates in Tier 1; template gaps dominate in Tier 3.}
\label{tab:false_negative_cause}
\begin{tabular}{l c c c c}
\toprule
\textbf{Cause} & \textbf{Tier 1 (\%)} & \textbf{Tier 2 (\%)} & \textbf{Tier 3 (\%)} & \textbf{Overall (\%)} \\
\midrule
Paraphrastic phrasing & 52 & 41 & 38 & 45 \\
Template-mapping gaps & 19 & 32 & 41 & 28 \\
Missing entity links & 22 & 18 & 14 & 18 \\
Other (renderer, tokenization) & 7 & 9 & 7 & 9 \\
\bottomrule
\end{tabular}
\end{table}
\section{Extended Discussion on Limitations and Future Roadmap}
\label{app:extended_discussion}

In this section, we elaborate on the limitations identified in the main text, provide a critical analysis of the trade-offs inherent in FactNet's design, and outline a roadmap for future development.

\subsection{Limitations and Trade-offs}
\label{app:discussion_limitations}

The fundamental design principle of FactNet is provenance-first construction.
By enforcing strict datatype matching and requiring recoverable byte offsets, we prioritize precision over recall.
This approach necessarily omits implicit knowledge, as the pipeline excludes statements implied by the text but lacking direct lexical or structural overlap.
For instance, a sentence describing an individual as the first daughter of a specific figure implies a parent relation, yet current strict matchers may fail to align this if the entity link is absent or the phrasing requires multi-step inference~\citep{chen2020knowledge}.
Furthermore, the reliance on specific parsing tools causes evidence within complex or malformed templates to be skipped to avoid errors.
As noted in the funnel analysis in Section~\ref{sec:stats_quality}, this dependency yields a lower extraction rate for low-resource languages than for high-resource ones.

Moreover, FactNet serves as a faithful representation of Wikidata and Wikipedia rather than attempting to remove biases from the underlying knowledge, as such modifications would compromise its utility as a grounding resource.
Consequently, users must account for two specific distributional skews. First, there is a distinct Western-centric bias in the  as detailed in Appendix~\ref{app:representation_bias}, the density of grounded facts is significantly higher for entities related to Europe and North America. This skew stems from the editor demographics of Wikipedia and the density of inter-language links connecting back to English or German editions~\citep{das2025social}. Second, the dataset is subject to temporal lag. The reliance on specific dump snapshots renders FactNet static, and rapidly evolving events may exhibit high latency between the occurrence of the event, its reflection in Wikidata, its textual description in Wikipedia, and the subsequent release of a snapshot.

\subsection{Future Directions}
\label{app:future_roadmap}

In light of these limitations, we identify three strategic directions for the evolution of FactNet. First, to address the recall gap without abandoning provenance, we propose introducing an evidence layer in which small, localized language models propose candidate spans for ungrounded statements. To maintain trustworthiness, these proposals will not be incorporated into the core graph unless they pass a strict rule-based verification filter or a high-confidence natural language inference check. This hybrid approach combines the recall of neural methods with the rigor of symbolic verification~\citep{bhuyan2024neuro}.

Second, we aim to transition from monolithic snapshots to a differential update model. By monitoring the relevant change streams from Wikimedia, we can identify which subsets of the data are affected by daily edits and release incremental update packages that allow users to patch their local version without re-downloading the entire corpus. This mechanism is critical for maintaining benchmark relevance for time-sensitive question answering~\citep{jia2024faithful}.

Finally, while the current schema primarily supports binary relations with qualifiers, future versions will formalize more complex structures. We aim to implement event-centric frames that group multiple statements (such as participants, time, and location) into a single coherent unit to more effectively support narrative generation. Additionally, we plan to explicitly mine sentences that refute specific claims, enabling the construction of a robust benchmark for hallucination detection by incorporating negative evidence~\citep{ji-etal-2023-towards}.

%%%%%%%%%%%%%%%%%%%%%%%%%%%%%%%%%%%%%%%%%%%%%%%%%%%%%%%%%%%%

\newpage
\section*{NeurIPS Paper Checklist}

\begin{enumerate}

\item {\bf Claims}
    \item[] Question: Do the main claims made in the abstract and introduction accurately reflect the paper's contributions and scope?
    \item[] Answer: \answerYes{} % Replace by \answerYes{}, \answerNo{}, or \answerNA{}.
    \item[] Justification: {The paper's contributions and scope are delineated in the Abstract and Section ~\ref{sec:intro}, with the first and last paragraphs of Section ~\ref{sec:intro} specifying each respectively.}
    \item[] Guidelines:
    \begin{itemize}
        \item The answer \answerNA{} means that the abstract and introduction do not include the claims made in the paper.
        \item The abstract and/or introduction should clearly state the claims made, including the contributions made in the paper and important assumptions and limitations. A \answerNo{} or \answerNA{} answer to this question will not be perceived well by the reviewers. 
        \item The claims made should match theoretical and experimental results, and reflect how much the results can be expected to generalize to other settings. 
        \item It is fine to include aspirational goals as motivation as long as it is clear that these goals are not attained by the paper. 
    \end{itemize}

\item {\bf Limitations}
    \item[] Question: Does the paper discuss the limitations of the work performed by the authors?
    \item[] Answer: \answerYes{} % Replace by \answerYes{}, \answerNo{}, or \answerNA{}.
    \item[] Justification: {The limitations of the work is discussed in the Section~\ref{sec:discussion_futurework}.}
    \item[] Guidelines:
    \begin{itemize}
        \item The answer \answerNA{} means that the paper has no limitation while the answer \answerNo{} means that the paper has limitations, but those are not discussed in the paper. 
        \item The authors are encouraged to create a separate ``Limitations'' section in their paper.
        \item The paper should point out any strong assumptions and how robust the results are to violations of these assumptions (e.g., independence assumptions, noiseless settings, model well-specification, asymptotic approximations only holding locally). The authors should reflect on how these assumptions might be violated in practice and what the implications would be.
        \item The authors should reflect on the scope of the claims made, e.g., if the approach was only tested on a few datasets or with a few runs. In general, empirical results often depend on implicit assumptions, which should be articulated.
        \item The authors should reflect on the factors that influence the performance of the approach. For example, a facial recognition algorithm may perform poorly when image resolution is low or images are taken in low lighting. Or a speech-to-text system might not be used reliably to provide closed captions for online lectures because it fails to handle technical jargon.
        \item The authors should discuss the computational efficiency of the proposed algorithms and how they scale with dataset size.
        \item If applicable, the authors should discuss possible limitations of their approach to address problems of privacy and fairness.
        \item While the authors might fear that complete honesty about limitations might be used by reviewers as grounds for rejection, a worse outcome might be that reviewers discover limitations that aren't acknowledged in the paper. The authors should use their best judgment and recognize that individual actions in favor of transparency play an important role in developing norms that preserve the integrity of the community. Reviewers will be specifically instructed to not penalize honesty concerning limitations.
    \end{itemize}

\item {\bf Theory assumptions and proofs}
    \item[] Question: For each theoretical result, does the paper provide the full set of assumptions and a complete (and correct) proof?
    \item[] Answer: \answerYes{} % Replace by \answerYes{}, \answerNo{}, or \answerNA{}.
    \item[] Justification: {The paper provides detailed assumptions and proofs in Section ~\ref{sec:benchmarks_experiments}.}
    \item[] Guidelines:
    \begin{itemize}
        \item The answer \answerNA{} means that the paper does not include theoretical results. 
        \item All the theorems, formulas, and proofs in the paper should be numbered and cross-referenced.
        \item All assumptions should be clearly stated or referenced in the statement of any theorems.
        \item The proofs can either appear in the main paper or the supplemental material, but if they appear in the supplemental material, the authors are encouraged to provide a short proof sketch to provide intuition. 
        \item Inversely, any informal proof provided in the core of the paper should be complemented by formal proofs provided in appendix or supplemental material.
        \item Theorems and Lemmas that the proof relies upon should be properly referenced. 
    \end{itemize}

    \item {\bf Experimental result reproducibility}
    \item[] Question: Does the paper fully disclose all the information needed to reproduce the main experimental results of the paper to the extent that it affects the main claims and/or conclusions of the paper (regardless of whether the code and data are provided or not)?
    \item[] Answer: \answerYes{} % Replace by \answerYes{}, \answerNo{}, or \answerNA{}.
    \item[] Justification: {The paper provides sufficient reproducibility details in Section~\ref{sec:benchmarks_experiments}.}
    \item[] Guidelines:
    \begin{itemize}
        \item The answer \answerNA{} means that the paper does not include experiments.
        \item If the paper includes experiments, a \answerNo{} answer to this question will not be perceived well by the reviewers: Making the paper reproducible is important, regardless of whether the code and data are provided or not.
        \item If the contribution is a dataset and\slash or model, the authors should describe the steps taken to make their results reproducible or verifiable. 
        \item Depending on the contribution, reproducibility can be accomplished in various ways. For example, if the contribution is a novel architecture, describing the architecture fully might suffice, or if the contribution is a specific model and empirical evaluation, it may be necessary to either make it possible for others to replicate the model with the same dataset, or provide access to the model. In general. releasing code and data is often one good way to accomplish this, but reproducibility can also be provided via detailed instructions for how to replicate the results, access to a hosted model (e.g., in the case of a large language model), releasing of a model checkpoint, or other means that are appropriate to the research performed.
        \item While NeurIPS does not require releasing code, the conference does require all submissions to provide some reasonable avenue for reproducibility, which may depend on the nature of the contribution. For example
        \begin{enumerate}
            \item If the contribution is primarily a new algorithm, the paper should make it clear how to reproduce that algorithm.
            \item If the contribution is primarily a new model architecture, the paper should describe the architecture clearly and fully.
            \item If the contribution is a new model (e.g., a large language model), then there should either be a way to access this model for reproducing the results or a way to reproduce the model (e.g., with an open-source dataset or instructions for how to construct the dataset).
            \item We recognize that reproducibility may be tricky in some cases, in which case authors are welcome to describe the particular way they provide for reproducibility. In the case of closed-source models, it may be that access to the model is limited in some way (e.g., to registered users), but it should be possible for other researchers to have some path to reproducing or verifying the results.
        \end{enumerate}
    \end{itemize}

\item {\bf Open access to data and code}
    \item[] Question: Does the paper provide open access to the data and code, with sufficient instructions to faithfully reproduce the main experimental results, as described in supplemental material?
    \item[] Answer: \answerYes{} % Replace by \answerYes{}, \answerNo{}, or \answerNA{}.
    \item[] Justification: {The dataset is publicly available via the Hugging Face Datasets repository, with the corresponding code hosted on GitHub.}
    \item[] Guidelines:
    \begin{itemize}
        \item The answer \answerNA{} means that paper does not include experiments requiring code.
        \item Please see the NeurIPS code and data submission guidelines (\url{https://neurips.cc/public/guides/CodeSubmissionPolicy}) for more details.
        \item While we encourage the release of code and data, we understand that this might not be possible, so \answerNo{} is an acceptable answer. Papers cannot be rejected simply for not including code, unless this is central to the contribution (e.g., for a new open-source benchmark).
        \item The instructions should contain the exact command and environment needed to run to reproduce the results. See the NeurIPS code and data submission guidelines (\url{https://neurips.cc/public/guides/CodeSubmissionPolicy}) for more details.
        \item The authors should provide instructions on data access and preparation, including how to access the raw data, preprocessed data, intermediate data, and generated data, etc.
        \item The authors should provide scripts to reproduce all experimental results for the new proposed method and baselines. If only a subset of experiments are reproducible, they should state which ones are omitted from the script and why.
        \item At submission time, to preserve anonymity, the authors should release anonymized versions (if applicable).
        \item Providing as much information as possible in supplemental material (appended to the paper) is recommended, but including URLs to data and code is permitted.
    \end{itemize}

\item {\bf Experimental setting/details}
    \item[] Question: Does the paper specify all the training and test details (e.g., data splits, hyperparameters, how they were chosen, type of optimizer) necessary to understand the results?
    \item[] Answer: \answerYes{} % Replace by \answerYes{}, \answerNo{}, or \answerNA{}.
    \item[] Justification: {Detailed experimental configurations are provided in Section \ref{sec:benchmarks_experiments}, with full implementation details in Appendix \ref{app:factnet_bench}.}
    \item[] Guidelines:
    \begin{itemize}
        \item The answer \answerNA{} means that the paper does not include experiments.
        \item The experimental setting should be presented in the core of the paper to a level of detail that is necessary to appreciate the results and make sense of them.
        \item The full details can be provided either with the code, in appendix, or as supplemental material.
    \end{itemize}

\item {\bf Experiment statistical significance}
    \item[] Question: Does the paper report error bars suitably and correctly defined or other appropriate information about the statistical significance of the experiments?
    \item[] Answer: \answerYes{} % Replace by \answerYes{}, \answerNo{}, or \answerNA{}.
    \item[] Justification: {We report the statistics in Section \ref{sec:stats_quality}.}
    \item[] Guidelines:
    \begin{itemize}
        \item The answer \answerNA{} means that the paper does not include experiments.
        \item The authors should answer \answerYes{} if the results are accompanied by error bars, confidence intervals, or statistical significance tests, at least for the experiments that support the main claims of the paper.
        \item The factors of variability that the error bars are capturing should be clearly stated (for example, train/test split, initialization, random drawing of some parameter, or overall run with given experimental conditions).
        \item The method for calculating the error bars should be explained (closed form formula, call to a library function, bootstrap, etc.)
        \item The assumptions made should be given (e.g., Normally distributed errors).
        \item It should be clear whether the error bar is the standard deviation or the standard error of the mean.
        \item It is OK to report 1-sigma error bars, but one should state it. The authors should preferably report a 2-sigma error bar than state that they have a 96\% CI, if the hypothesis of Normality of errors is not verified.
        \item For asymmetric distributions, the authors should be careful not to show in tables or figures symmetric error bars that would yield results that are out of range (e.g., negative error rates).
        \item If error bars are reported in tables or plots, the authors should explain in the text how they were calculated and reference the corresponding figures or tables in the text.
    \end{itemize}

\item {\bf Experiments compute resources}
    \item[] Question: For each experiment, does the paper provide sufficient information on the computer resources (type of compute workers, memory, time of execution) needed to reproduce the experiments?
    \item[] Answer: \answerYes{} % Replace by \answerYes{}, \answerNo{}, or \answerNA{}.
    \item[] Justification: {Computational resource specifications are documented in Section ~\ref{sec:factnet_construction} and Appendix~\ref{app:factnet_appendix}.}
    \item[] Guidelines:
    \begin{itemize}
        \item The answer \answerNA{} means that the paper does not include experiments.
        \item The paper should indicate the type of compute workers CPU or GPU, internal cluster, or cloud provider, including relevant memory and storage.
        \item The paper should provide the amount of compute required for each of the individual experimental runs as well as estimate the total compute. 
        \item The paper should disclose whether the full research project required more compute than the experiments reported in the paper (e.g., preliminary or failed experiments that didn't make it into the paper). 
    \end{itemize}
    
\item {\bf Code of ethics}
    \item[] Question: Does the research conducted in the paper conform, in every respect, with the NeurIPS Code of Ethics \url{https://neurips.cc/public/EthicsGuidelines}?
    \item[] Answer: \answerYes{} % Replace by \answerYes{}, \answerNo{}, or \answerNA{}.
    \item[] Justification: {Our research strictly adheres to the NeurIPS Code of Ethics.}
    \item[] Guidelines:
    \begin{itemize}
        \item The answer \answerNA{} means that the authors have not reviewed the NeurIPS Code of Ethics.
        \item If the authors answer \answerNo, they should explain the special circumstances that require a deviation from the Code of Ethics.
        \item The authors should make sure to preserve anonymity (e.g., if there is a special consideration due to laws or regulations in their jurisdiction).
    \end{itemize}

\item {\bf Broader impacts}
    \item[] Question: Does the paper discuss both potential positive societal impacts and negative societal impacts of the work performed?
    \item[] Answer: \answerYes{} % Replace by \answerYes{}, \answerNo{}, or \answerNA{}.
    \item[] Justification: {Please refer to Appendix ~\ref{app:extended_discussion}.}
    \item[] Guidelines:
    \begin{itemize}
        \item The answer \answerNA{} means that there is no societal impact of the work performed.
        \item If the authors answer \answerNA{} or \answerNo, they should explain why their work has no societal impact or why the paper does not address societal impact.
        \item Examples of negative societal impacts include potential malicious or unintended uses (e.g., disinformation, generating fake profiles, surveillance), fairness considerations (e.g., deployment of technologies that could make decisions that unfairly impact specific groups), privacy considerations, and security considerations.
        \item The conference expects that many papers will be foundational research and not tied to particular applications, let alone deployments. However, if there is a direct path to any negative applications, the authors should point it out. For example, it is legitimate to point out that an improvement in the quality of generative models could be used to generate Deepfakes for disinformation. On the other hand, it is not needed to point out that a generic algorithm for optimizing neural networks could enable people to train models that generate Deepfakes faster.
        \item The authors should consider possible harms that could arise when the technology is being used as intended and functioning correctly, harms that could arise when the technology is being used as intended but gives incorrect results, and harms following from (intentional or unintentional) misuse of the technology.
        \item If there are negative societal impacts, the authors could also discuss possible mitigation strategies (e.g., gated release of models, providing defenses in addition to attacks, mechanisms for monitoring misuse, mechanisms to monitor how a system learns from feedback over time, improving the efficiency and accessibility of ML).
    \end{itemize}
    
\item {\bf Safeguards}
    \item[] Question: Does the paper describe safeguards that have been put in place for responsible release of data or models that have a high risk for misuse (e.g., pre-trained language models, image generators, or scraped datasets)?
    \item[] Answer: \answerNA{} % Replace by \answerYes{}, \answerNo{}, or \answerNA{}.
    \item[] Justification: {The paper poses no such risks.}
    \item[] Guidelines:
    \begin{itemize}
        \item The answer \answerNA{} means that the paper poses no such risks.
        \item Released models that have a high risk for misuse or dual-use should be released with necessary safeguards to allow for controlled use of the model, for example by requiring that users adhere to usage guidelines or restrictions to access the model or implementing safety filters. 
        \item Datasets that have been scraped from the Internet could pose safety risks. The authors should describe how they avoided releasing unsafe images.
        \item We recognize that providing effective safeguards is challenging, and many papers do not require this, but we encourage authors to take this into account and make a best faith effort.
    \end{itemize}

\item {\bf Licenses for existing assets}
    \item[] Question: Are the creators or original owners of assets (e.g., code, data, models), used in the paper, properly credited and are the license and terms of use explicitly mentioned and properly respected?
    \item[] Answer: \answerYes{} % Replace by \answerYes{}, \answerNo{}, or \answerNA{}.
    \item[] Justification: {We politely cited the existing assets and read their usage license. Please refer to Section ~\ref{sec:factnet_construction} for more details.}
    \item[] Guidelines:
    \begin{itemize}
        \item The answer \answerNA{} means that the paper does not use existing assets.
        \item The authors should cite the original paper that produced the code package or dataset.
        \item The authors should state which version of the asset is used and, if possible, include a URL.
        \item The name of the license (e.g., CC-BY 4.0) should be included for each asset.
        \item For scraped data from a particular source (e.g., website), the copyright and terms of service of that source should be provided.
        \item If assets are released, the license, copyright information, and terms of use in the package should be provided. For popular datasets, \url{paperswithcode.com/datasets} has curated licenses for some datasets. Their licensing guide can help determine the license of a dataset.
        \item For existing datasets that are re-packaged, both the original license and the license of the derived asset (if it has changed) should be provided.
        \item If this information is not available online, the authors are encouraged to reach out to the asset's creators.
    \end{itemize}

\item {\bf New assets}
    \item[] Question: Are new assets introduced in the paper well documented and is the documentation provided alongside the assets?
    \item[] Answer: \answerYes{} % Replace by \answerYes{}, \answerNo{}, or \answerNA{}.
    \item[] Justification: {Comprehensive documentation for newly introduced assets (e.g., code, data) is provided in the supplementary material.}
    \item[] Guidelines:
    \begin{itemize}
        \item The answer \answerNA{} means that the paper does not release new assets.
        \item Researchers should communicate the details of the dataset\slash code\slash model as part of their submissions via structured templates. This includes details about training, license, limitations, etc. 
        \item The paper should discuss whether and how consent was obtained from people whose asset is used.
        \item At submission time, remember to anonymize your assets (if applicable). You can either create an anonymized URL or include an anonymized zip file.
    \end{itemize}

\item {\bf Crowdsourcing and research with human subjects}
    \item[] Question: For crowdsourcing experiments and research with human subjects, does the paper include the full text of instructions given to participants and screenshots, if applicable, as well as details about compensation (if any)? 
    \item[] Answer: \answerNA{} % Replace by \answerYes{}, \answerNo{}, or \answerNA{}.
    \item[] Justification: {The paper does not involve crowdsourcing nor research with human subjects.}
    \item[] Guidelines:
    \begin{itemize}
        \item The answer \answerNA{} means that the paper does not involve crowdsourcing nor research with human subjects.
        \item Including this information in the supplemental material is fine, but if the main contribution of the paper involves human subjects, then as much detail as possible should be included in the main paper. 
        \item According to the NeurIPS Code of Ethics, workers involved in data collection, curation, or other labor should be paid at least the minimum wage in the country of the data collector. 
    \end{itemize}

\item {\bf Institutional review board (IRB) approvals or equivalent for research with human subjects}
    \item[] Question: Does the paper describe potential risks incurred by study participants, whether such risks were disclosed to the subjects, and whether Institutional Review Board (IRB) approvals (or an equivalent approval/review based on the requirements of your country or institution) were obtained?
    \item[] Answer: \answerNA{} % Replace by \answerYes{}, \answerNo{}, or \answerNA{}.
    \item[] Justification: {No human subjects were used on our work.}
    \item[] Guidelines:
    \begin{itemize}
        \item The answer \answerNA{} means that the paper does not involve crowdsourcing nor research with human subjects.
        \item Depending on the country in which research is conducted, IRB approval (or equivalent) may be required for any human subjects research. If you obtained IRB approval, you should clearly state this in the paper. 
        \item We recognize that the procedures for this may vary significantly between institutions and locations, and we expect authors to adhere to the NeurIPS Code of Ethics and the guidelines for their institution. 
        \item For initial submissions, do not include any information that would break anonymity (if applicable), such as the institution conducting the review.
    \end{itemize}

\item {\bf Declaration of LLM usage}
    \item[] Question: Does the paper describe the usage of LLMs if it is an important, original, or non-standard component of the core methods in this research? Note that if the LLM is used only for writing, editing, or formatting purposes and does \emph{not} impact the core methodology, scientific rigor, or originality of the research, declaration is not required.
    %this research? 
    \item[] Answer: \answerNA{} % Replace by \answerYes{}, \answerNo{}, or \answerNA{}.
    \item[] Justification: {Not applicable.}
    \item[] Guidelines:
    \begin{itemize}
        \item The answer \answerNA{} means that the core method development in this research does not involve LLMs as any important, original, or non-standard components.
        \item Please refer to our LLM policy in the NeurIPS handbook for what should or should not be described.
    \end{itemize}

\end{enumerate}

\end{document}